\pdfoutput=1

\documentclass[11pt]{article}

\usepackage{acl}

\usepackage{times}
\usepackage{latexsym}
\usepackage{enumitem}
\usepackage[T1]{fontenc}

\usepackage[utf8]{inputenc}

\usepackage{microtype}

\usepackage{inconsolata}

\usepackage{hyperref}
\usepackage{url}
\usepackage[inkscapelatex=false]{svg}
\usepackage{tabularx}
\usepackage{multirow}
\usepackage{booktabs}
\usepackage{caption}
\usepackage{graphicx}
\usepackage{amsmath}
\usepackage[capitalise,noabbrev]{cleveref}
\usepackage{subcaption}
\usepackage{xcolor}
\usepackage{makecell}
\usepackage{rotating}

\newcommand{\ourdataabbr}{TAQA}

\definecolor{darkred}{RGB}{200,0,0}

\title{Set the Clock: Temporal Alignment of Pretrained Language Models}

\newcommand*\samethanks[1][\value{footnote}]{\footnotemark[#1]}
\author{
    Bowen Zhao\thanks{Equal contribution.}$^{\spadesuit}$ \; \;
    Zander Brumbaugh\samethanks{}$^{\spadesuit}$ \; \;
    Yizhong Wang\samethanks{}$^{\spadesuit\clubsuit}$ \; \; \\
    \textbf{Hannaneh Hajishirzi$^{\spadesuit\clubsuit}$\; \; \; 
    Noah A. Smith$^{\spadesuit\clubsuit}$ 
    } \\ \\
    $^\spadesuit$University of Washington \;
    $^\clubsuit$Allen Institute for AI \\
    \texttt{bowen98@uw.edu \; \{brumbzan,yizhongw\}@cs.washington.edu}
}

\begin{document}
\maketitle

\begin{abstract}

Language models (LMs) are trained on web text originating from many points in time and, in general, without any explicit temporal grounding. 
This work investigates the {\it temporal chaos} of pretrained LMs and explores various methods to align their internal knowledge to a target time, which we call ``temporal alignment.'' 
To do this, we first automatically construct a dataset containing 20K time-sensitive questions and their answers for each year from 2000 to 2023. Based on this dataset, we empirically show that pretrained LMs (e.g., LLaMa2), despite having a recent pretraining cutoff (e.g., 2022), mostly answer questions using earlier knowledge (e.g., in 2019). 
We then develop several methods, from prompting to finetuning, to align LMs to use their most recent knowledge when answering questions, and investigate various factors in this alignment.
Our experiments demonstrate that aligning LLaMa2 to the year 2022 can enhance its performance by up to 62\% according to that year's answers. This improvement occurs even without explicitly mentioning time information, indicating the possibility of aligning models' internal sense of time after pretraining.
Finally, we find that alignment to a historical time is also possible, with up to $2.8\times$ the performance of the unaligned LM in 2010 if finetuning models to that year. 
These findings hint at 
the sophistication of LMs\textquotesingle~ internal knowledge organization and the necessity of tuning them properly.\footnote{Our dataset and code is available at \url{https://github.com/yizhongw/llm-temporal-alignment}.}

\end{abstract}

\section{Introduction}
\label{sec:intro}

\begin{figure*}[t]
    \centering
    \includegraphics[width=\textwidth, trim=0 1cm 0 2cm]{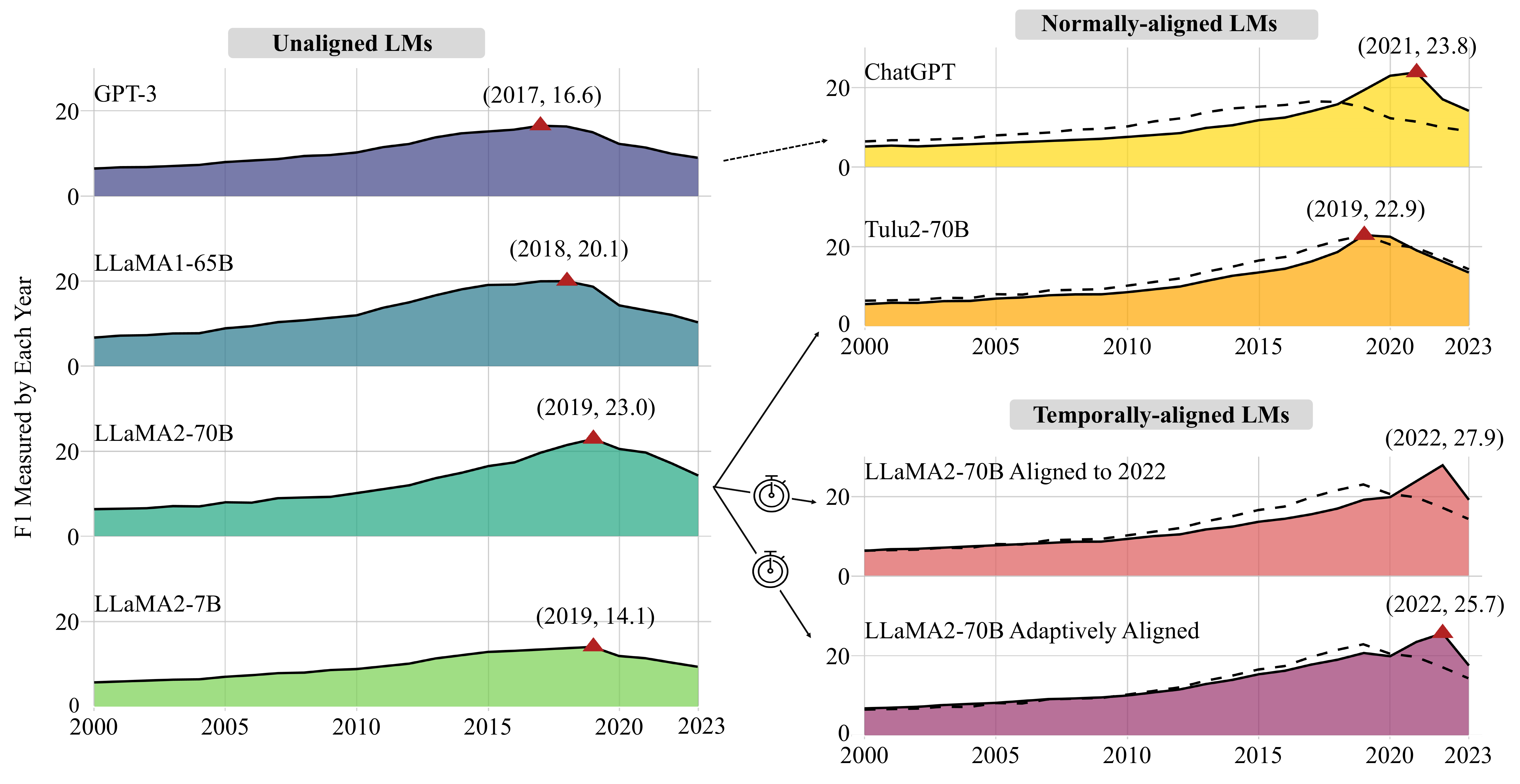}
    \caption{
    Performance (F1 score) of various LMs on our \ourdataabbr{} dataset, by year. 
    Unaligned LMs (left) and conventionally aligned models (upper right) show relatively stronger performance when measured by the answers in earlier years, with their predictions more scattered across time.
    Our temporal alignment methods (lower right) lead to improved performance closer to a recent time (here, 2022) with a higher peak. The dotted line between GPT-3 and ChatGPT implies an uncertain relation (the latter is not necessarily derived from the former).
 }
    \label{fig:teaser}
\end{figure*}

Large-scale pretraining \citep[i.a.]{devlin2019bert, raffel2020exploring, brown2020language} has enabled language models (LMs) to learn extensive knowledge from unlabeled text~\citep{petroni-etal-2019-language}. However, since pretraining corpora are constructed over a wide time period, they 
inevitably contain outdated and contradictory information \citep{longpre2023pretrainers}. 
Previous studies have found that the temporal misalignment between LM pretraining and deployment has a significant impact on models' performance \citep{NEURIPS2021_f5bf0ba0, agarwal2022temporal, luu2022time}, which motivates many studies on making model's knowledge up-to-date, by either continual learning \citep{jin-etal-2022-lifelong-pretraining,ke2023continual}, knowledge editing ~\citep{mitchell2022fast,meng2023massediting}, or retrieval augmentation \citep{zhang2023mitigating,vu2023freshllms}.
These approaches mainly focus on updating models with new knowledge and do not evaluate LM's internal temporal knowledge across time. %

In this work, we hypothesize that LMs, after pretraining, encode a chaotic sense of time, which means that models do not know which time-sensitive knowledge to use even if they have seen it during pretraining. We empirically investigate this temporal chaos and explore methods to do {\it temporal alignment}, which aims to align models' internal knowledge to a target time.

We first introduce our TemporalAlignmentQA dataset (\ourdataabbr{}, \S\ref{sec:data}) consisting of 20,148 questions, each with at least five different answers between 2000 and 2023. This dataset is automatically constructed based on Wikipedia tables that contain temporal information, making it easily scalable for future study as the world keeps changing. Complementary to previous temporal QA datasets \citep{chen2021dataset,dhingra2022time,kasai2022realtime, liska2022streamingqa,tan2023towards,wei2023menatqa}, \ourdataabbr{} focuses on facts that changed multiple times during a recent period when most pretraining data is collected, so that it can be used to probe the temporal knowledge distribution of pretrained LMs. \autoref{fig:teaser} plots the F1 score of several representative LMs relative to ground-truth answers from each year, demonstrating that they tend to use earlier knowledge to answer questions, even when they have a very recent pretraining cutoff date (e.g., performance peaks in 2019 for LLaMa2, which has a cutoff of September 2022; \citealp{touvron2023llama2}).

Next, we explore how to align LMs to answer questions based on a target time. Specifically, we first try aligning models to the most recent possible time, as given by their knowledge cutoff date (\S\ref{sec:recency_alignment}). This is often desired when LMs are deployed for general user populations who seek current answers to their questions. We propose three methods, including 1) \textit{time-aware prompting}, where we prompt LMs using time information and time-sensitive examples; 2) \textit{target-year finetuning}, where we finetune LMs with ground truth answers in a target year so that they can generalize and answer new questions based on that year; 3) \textit{adaptive finetuning}, where we elicit the most recent knowledge each LM knows about a question (which can be earlier if the model does not know the updated version), and teach the model to answer adaptively based on the year proper for its own.

Experimental results (\S\ref{sec:experiments}) show that finetuning LLaMa2 on {\ourdataabbr}, even without explicit temporal information, can relatively improve the answer performance in 2022 by up to 62.2\%. 
We additionally explore the possibility of aligning LMs to a historical time (\S\ref{sec:align-to-historical}) and find our finetuning alignment strategy can boost the QA performance by $2.8\times$ when aligning to 2010. We also find that 2019 is the most readily alignable year for LLaMa2, while aligning it to 2015 causes the fewest errors measured against all valid answers from 2000 to 2023.

\begin{figure*}[t!]
    \centering
    \includegraphics[width=0.95\textwidth, trim=0 0.5cm 0 2cm]{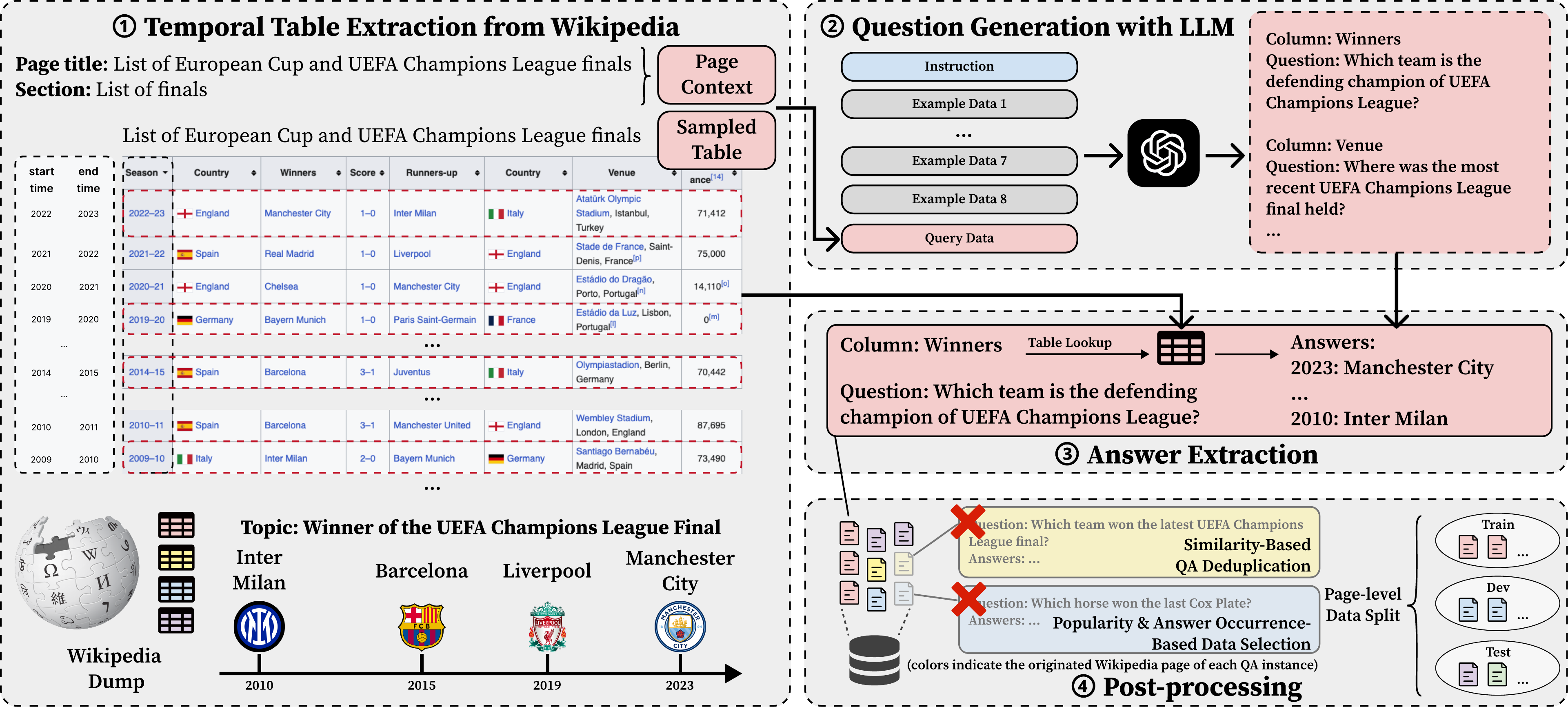}
    \caption{The data construction process of our {\ourdataabbr} dataset.}
    \label{fig:data-construction}
\end{figure*}

\section{\ourdataabbr: A QA Dataset for Studying LM Temporal Alignment}
\label{sec:data}
To study the temporal chaos of pretrained LMs and how to align them temporally,
we curate the \ourdataabbr{} dataset, featuring questions with frequently changing answers in a recent time period (2000--2023). 
We introduce the formulation of {\ourdataabbr} in \S\ref{subsec:formulation}, detail the data construction process in \S\ref{subsec:data-construction}, and further describe the evaluation metrics in \S\ref{subsec:metrics}.

\subsection{Problem Formulation} \label{subsec:formulation}
We formally define the task of \ourdataabbr{} as follows: given a question $q$, which has a set of answers $A = \{(a, t_s, t_e)\}$, where each answer $a$ is only correct when the question is being asked during times $t \in [t^s, t^e]$. Such a dataset can be represented as $\mathcal{D} = \{(q_i, A_i)\}_{i=1}^N$, where at a specific timepoint $t$, the valid answers of question $q$ shall be $\mathcal{A}(q, t) =$
\begin{equation}
    \small
     \{ a \mid \exists (a, t^s, t^e) \in A \wedge (q, A) \in \mathcal{D} \text{, s.t. } t \in [t^s, t^e]  \}
\end{equation}
We consider a response from a QA system $\hat{a}$ to question $q$ at time $t$ as correct as long as $\hat{a} \in \mathcal{A}(q, t)$. Since only questions with varied answers through history are useful to us, we quantify a question's time-sensitivity as the number of unique answers during a time, i.e., 
\begin{equation}
    \small
    \mathcal{S}(q, t_s, t_e) = |\{\mathcal{A}(q, t)\mid t\in [t_s, t_e]\}| \label{eq:sensitivity}
\end{equation}
In practice, we only keep questions such that $\mathcal{S}(q, 2000, 2023) \ge 5\}$ in our \ourdataabbr{} dataset.

\subsection{Data Construction} \label{subsec:data-construction}
To get question $q$ with its multiple temporal answers $(a, t_s, t_e)$, we use tables with temporal columns from the Wikipedia dump to build our {\ourdataabbr} dataset, as the temporal columns can be seen as qualifiers $t_s, t_e$ and other columns can be leveraged to generate QA pairs.
The advantage of Wikipedia is that its topics are popular and well-structured. Also, the topic distribution of Wikipedia is less biased than open knowledge graphs, such as Wikidata~\citep{10.1145/3125433.3125445}. Furthermore, Wikipedia is used in the pretraining corpus for most modern LMs, ensuring that LMs have learned such knowledge from pretraining. We demonstrate our steps for constructing the \ourdataabbr{} dataset as shown in \cref{fig:data-construction} as follows.

\paragraph{Temporal table extraction from Wikipedia.} Given the Wikipedia dump,\footnote{We use the English Wikipedia dump of January 1, 2024.} we use the WikiExtractor~\citep{Wikiextractor2015} to extract all tables from the dump into the CSV format. Then, we use heuristic-based methods to identify the columns denoting temporal information (i.e., $(t_s, t_e)$) for answers. As we want to gather questions for contemporary knowledge only, we only keep tables with information for every year from 2010 to 2023, resulting in 17,932 tables.

\paragraph{Question generation with LMs.} To generate natural questions with varied styles, we prompt GPT-4 with our manually curated few-shot examples with the tables gathered. We also add each table's Wikipedia page information (title and section names) to the prompt to provide sufficient context.
To reduce the query cost, we sample the table and use only the rows corresponding to 2010, 2020, and 2023. To extract answers from the table later, we instruct GPT-4 to generate the column name before each question. After this process, 96,309 question-column pairs are generated in total. Appendix \S\ref{sec:qg_details} presents the detailed prompt.

\paragraph{Answer extraction.} With the question-column pairs generated, we use the column names to extract the answers to curate QA pairs from the table directly. When multiple rows have a shared period $[t_s, t_e]$, we treat all answers in those rows as valid. To ensure the question $q$ aligns with our sensitivity requirement $\mathcal{S}(q, 2000, 2023) \ge 5$, we only keep the QA pairs with at least five unique answers.

\paragraph{QA postprocessing.} To avoid data leakage in our {\ourdataabbr} dataset, we discard QA instances based on their question and answer similarities, measured using BM 25 scores calculated with \citet{Lin_etal_SIGIR2021_Pyserini}. When duplicates are found, we keep the instance with fewer words in the question as they are often more natural and popular among real people.
Next, we attempt to reduce the dataset's answer bias as we do not want LMs to improve their QA performance by memorizing frequently occurring answers. We begin by identifying questions with answers that are numerical or occur more than 300 times and randomly sample only 10\% of them to be kept in our {\ourdataabbr} dataset. 

\begin{table}[t]
\setlength\tabcolsep{2pt}
\centering
\small
\begin{tabular}{@{}lrrrrr@{}}
\toprule
Split $\downarrow$ & \# Questions & \# Tables & \# Pages & Popularity & Sensitivity \\ \midrule
Train & 10,148     & 6,109   & 5,751      & 10912.7         & 14.6             \\
Dev.   & 1,000      & 582     & 539        & 5592.2          & 16.6             \\
Test  & 9,000      & 5,290   & 4,725      & 6812.5          & 16.9             \\ \bottomrule
\end{tabular}
\caption{Statistics of {\ourdataabbr} dataset. We use Wikipedia pages' averaged monthly pageview to measure questions' popularity, and sensitivity is defined as \autoref{eq:sensitivity}. Popularity and sensitivity are averages across examples in each split. 
}
\label{tab:data-characteristics}
\end{table}

\paragraph{Train-test split.} Finally, we randomly split the 20,148 questions into train, development, and test sets in {\ourdataabbr}. To better probe LMs' understanding of temporal knowledge like \cref{fig:teaser}, we only put questions that can be answered since 2000 into the dev.~and test sets. We also ensure that there is no QA-originated Wikipedia page overlap between the train, dev., and test sets, as we want to avoid LMs improving their QA performance by memorizing the training data's knowledge. The data characteristics are illustrated in \cref{tab:data-characteristics}.

\paragraph{Quality control.} To examine the overall quality of the generated TAQA dataset, we randomly sampled 200 questions along with their answers and manually verified their correctness. Among the 200 examples, 185 (92.5\%) instances are considered correct, while the incorrectly generated questions are mostly because tables without temporal information are used for QA generation (6/200, 3.0\%). Another major issue is that GPT-4 sometimes automatically adds conditional clauses to the question, thus making them not time-sensitive anymore (5/200, 2.5\%). 

\begin{table*}[htbp]
\centering
\small
\begin{tabular}{@{}lrrrrrr@{}}
\toprule
Dataset    & \# Questions & \begin{tabular}[c]{@{}r@{}}\# Post-2019\\ answer change\end{tabular} & Time  & Median year & \begin{tabular}[c]{@{}r@{}}Track\\ length\end{tabular} & \begin{tabular}[c]{@{}r@{}}Avg.\\ sensitivity\end{tabular} \\ \midrule
SituatedQA~\citep{zhang-choi-2021-situatedqa} & 5,061       & 1,523                                                               & 1270--2021   & 2005        & 8.5                                                         & 1.7                                                        \\
Templama~\citep{dhingra-etal-2022-time}   & 5,798       & 1,082                                                               & 2010--2020   & 2015        & 8.9                                                         & 3.3                                                        \\
TSQA~\citep{chen2021dataset}       & 5,061       & 134                                                                 & 1--2021      & 1963        & 26.0                                                        & 4.4                                                        \\
RealtimeQA~\citep{kasai2022realtime} & 5,152       & 0                                                                   & 2020--2024   & 2022        & 1.0                                                         & 1.0                                                        \\ \midrule
TAQA (ours)       & 20,148      & 19,316                                                              & 2000--2023   & 2013        & 21.5                                                        & 15.6                                                       \\ \bottomrule
\end{tabular}
\caption{Detailed dataset characteristics comparison between TAQA and existing time-sensitive QA datasets. Track length is defined as the average number of years a question's answer is recorded.}
\label{tab:dataset-comparison}
\end{table*}

\begin{figure}
    \centering
    \includegraphics[width=0.9\columnwidth]{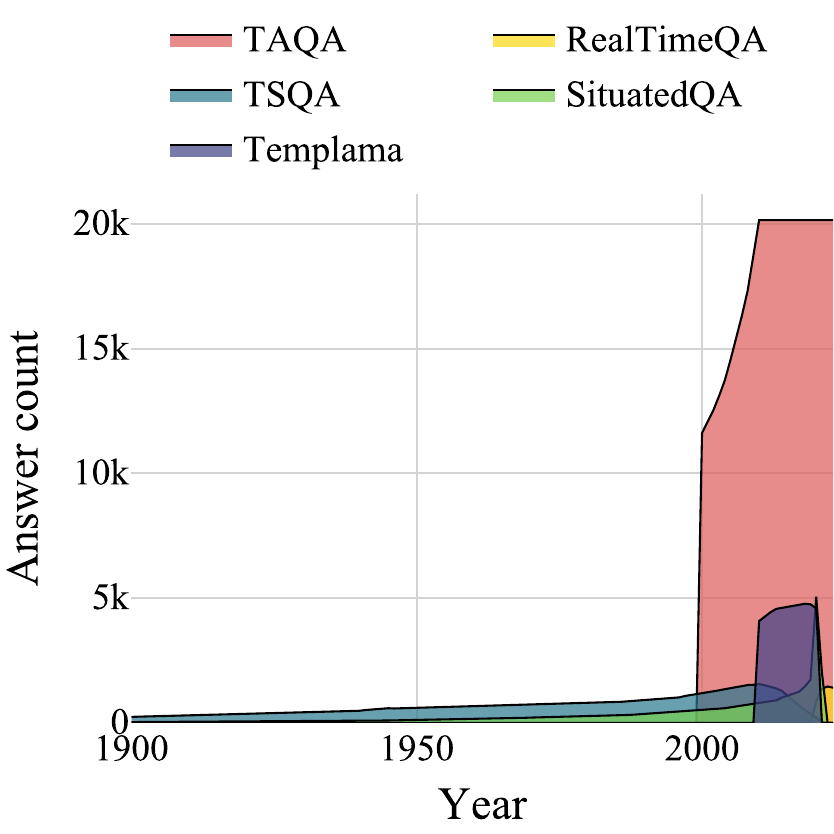}
    \caption{The comparison of TAQA and existing time-sensitive QA datasets regarding the frequency of the answers' corresponding time. TAQA contains more  data in the post-2000 era.}
    \label{fig:dataset-comparison}
\end{figure}

\paragraph{Dataset characteristics}
In \cref{tab:dataset-comparison}, we show   a detailed  comparison between TAQA and existing time-sensitive QA datasets. Overall, TAQA maintains more time-sensitive questions than existing datasets, making it sufficient for training LMs. Furthermore, questions in TAQA are more sensitive to time than existing datasets, and most answers to questions (95.9\%) were changed after 2019 when modern LMs were trained. TAQA also tracks time-sensitive questions over a long period of time (21.5 years on average). \cref{fig:dataset-comparison} also shows that TAQA contains more questions that have corresponding answers recorded after 2000\footnote{The Wikipedia tables we use and our intermediate data of TAQA contain answers before 2000 and after 2023, but we discard them for simplicity as these answers are not used in LM temporal alignment analysis.}. These properties make TAQA suitable for diagnosing modern LMs' understanding of time.

\subsection{Evaluation Metrics}
\label{subsec:metrics}

In \ourdataabbr, we focus on the question-answering task and expect models to output short answers. So, following prior QA evaluation \citep{rajpurkar2016squad, kwiatkowski2019natural}, we adopt the token-level F1 score to measure the quality of the LM-generated answers given ground truth answers. We further developed three variants tailored for the temporal alignment evaluation.

\paragraph{Target-year F1} Since the ground truth of a question $q$ in our {\ourdataabbr} dataset changes across time, we first calculate the F1 score of an LM-predicted answer $\hat{a}$ to the ground truth in each year between 2000 to 2023, for example, $\mathcal{F}^{2023}(\hat{a}) = \text{F1}(\hat{a}, a_{2023})$ for the LM-predicted answer's accuracy as of in 2023.

\paragraph{Max F1 over history} 
To determine whether an LM can answer questions with historically correct answers if the recent knowledge is unknown to the LM, we also calculate the maximum F1 score among different years' ground truth answers, i.e., $\displaystyle \mathcal{F}_{\text{max}}(\hat{a}) = \max_{2000 \leq i \leq 2023} \mathcal{F}^{i}(\hat{a})$. 

\paragraph{Decayed F1 towards target} Ideally, temporally aligned LMs' $\mathcal{F}_{\text{max}}$ scores should not be worse than the unaligned one. Furthermore, to differentiate the answers based on recent and outdated knowledge, we softly penalize the outdated answers by calculating the maximum value among F1 scores exponentially decayed by the time gap of the ground truths' year $i$ to the target year $j$, defined as:
\begin{equation}
    \small
    \mathcal{F}_{\text{decay}}^{j}(\hat{a}) = \max_{2000 \leq i \leq 2023} (\mathcal{F}^{i}(\hat{a}) \cdot \alpha^{|i-j|})
\end{equation}
where $0 < \alpha < 1$ is the decaying factor, which controls acceptance of the outdated answers. When $\alpha = 1$, $\mathcal{F}_{\text{decay}}^{j}$ is the same as $\mathcal{F}_{\text{max}}$, and when $\alpha \rightarrow 0$, $\mathcal{F}_{\text{decay}}^{j} \rightarrow \mathcal{F}^{j}$. In practice, we set $\alpha = 0.8$ so that a three-year-outdated answer will receive only half of its original F1 score.

\section{Aligning LMs towards Recency }
\label{sec:recency_alignment}

In this section, we propose several methods for aligning LMs to recent years. Complementary to existing studies that aim to teach language models up-to-date knowledge~\citep{jin-etal-2022-lifelong-pretraining,meng2023massediting}, our alignment methods focus on how to steer models to reorganize their knowledge and align them to the most-recent possible time.

\subsection{Time-Aware Prompting} \label{sec:prompting-alignment}
\label{subsec:time-aware-prompting}

An intuitive method to make LMs respond based on a specific time is to prompt them explicitly with the time information. We can achieve this by including the target year information. 
In addition, we also find models can benefit from being prompted with a few demonstration examples that are time-sensitive and with answers in the target time, in a sense that it is inferring the year information from these demonstration answers. We analyze the effects of these elements in Appendix \S\ref{sec:prompting_details} 
Therefore, we propose this time-aware prompting with the year information and a few time-sensitive examples in the prompt to activate models' knowledge of a specific year. Specifically, we append the prompt ``as of year $y$, the answer is'' at the end of each question $q$, where $y$ is the target year we want the LM to be aligned to. For the demonstration examples, we randomly sample 5 instances multiple times from the top-200 popular\footnote{Here, we use the QA's originated Wikipedia page's averaged monthly pageview from 2016 to 2023 to represent popularity.} QA instances in \ourdataabbr{} training set as few-shot examples, and choose the five examples that achieve the best performance on \ourdataabbr{} development set.

The chief advantage of this method is its simplicity, requiring no model updates; it can be used with models that only offer API access. However, prompting does not change models' internal state of time, and for many users, the additional information and few-shot examples are unnatural and add extra costs to inference.

\subsection{Target-Year Finetuning}
\label{subsec:target-year-finetuing}

We next explore ways to align LMs temporally by finetuning their parameters. As we want to change the models' internal sense of time, we choose not to add any temporal information in the context so that LMs can only generalize when they adjust parameters to model the time information. We refer readers to Appendix \S\ref{sec:training_details} for the training details.

One critical factor for finetuning is to select the data that can align LMs effectively. In fact, much up-to-date knowledge may not be presented to LMs during pretraining or not be memorized successfully. Tuning LMs with the knowledge they do not know can increase the risk of hallucination, 
i.e., LMs are encouraged to respond to questions with answers not seen during pretraining\footnote{This hallucination encouraging effect was mentioned in \url{https://news.berkeley.edu/2023/04/24/berkeley-talks-chatgpt-developer-john-schulman} and then widely discussed by the community.}. To address this, we propose selecting training data based on the correctness of sampled answers from the LMs. Given a question, we use the time-aware prompting in \S\ref{subsec:time-aware-prompting} to first sample ten outputs from the to-be-finetuned LM and calculate whether any of the ten samples can hit the ground truth answer. We select the top 5000 examples from the \ourdataabbr{} training set with the largest F1 overlaps between model samples and ground truth answers. By finetuning on question-answer pairs where the model already has some chance of giving the correct answer, we hope to steer it toward the target year rather than ``teaching it'' entirely new facts. We also explored other data selection methods based on popularity or model confidence, which are reported in \S\ref{subsec:data-selection}

\subsection{Temporal-Adaptive Finetuning} 
\label{subsec:temporal-adaptive}

Finetuning to a year assumes the model should answer all questions based on that target time; it, however, fails to take into account a desired graceful degradation:  when the LM doesn't have access to the target-year answer, it should revert back to the most recent prior answer.

To achieve this goal, we propose a temporally adaptive finetuning technique, where we dynamically determine the most recent and proper target year for each question. In practice, given a question, we iterate from the pretraining cutoff year (2022 for LLaMa2) to earlier years; for each year, we try using the correctness-based approach described in \S\ref{subsec:target-year-finetuing} to sample ten outputs, compare with the ground truth answers; if the F1 overlap is larger than a certain threshold (0.7), we will use this year as the target year for this question. As a result, we can adaptively assign each question with its expected target year, and then train the model to output ``Based on my latest knowledge for this question from year \dots, the answer is:'' before answering the question. Hence, we hope that the LM can learn to pick the target year for each question based on internal knowledge.

\section{Experiments}
\label{sec:experiments}

\begin{table*}[t!]
\centering
\small
\begin{tabular}{@{}p{0.17\textwidth}>{\centering}p{0.14\textwidth}ccccc@{}}
\toprule
Model $\downarrow$  &   Training Cutoff $\downarrow$    & $\mathcal{F}^{2022}$                        & $\mathcal{F}^{2022}_{\text{decay}}$        & $\mathcal{F}^{2021}$                       & $\mathcal{F}^{2021}_{\text{decay}}$        & $\mathcal{F}_{\text{max}}$                 \\ \midrule
LLaMA1-65B          & Aug. 2022                           & 12.1                                        & 26.8                                       & 13.2                                       & 29.5                                       & 54.5                                       \\
GPT-3          & Sep. 2021                           & 10.0                                        & 22.9                                       & 11.4                                       & 25.1                                       & 46.5                                       \\ \midrule
ChatGPT             & Sep. 2021                           & 17.1                                        & 32.5                                       & 23.8                                       & 35.5                                       & 46.0                                       \\
T\"ulu2-70B              & -                            & 16.2                                        & 32.2                                       & 19.1                                       & 35.1                                       & 50.1                                       \\ \midrule
LLaMA2-70B         & Sep. 2022                            & 17.2                                        & 34.1                                       & 19.8                                       & 37.0                                       & 55.9                                       \\
\multicolumn{2}{@{}l}{+ prompting to 2021  }    & 19.7 (\textcolor{darkblue}{+2.5})           & 38.9 (\textcolor{darkblue}{+4.8})          & 25.9 (\textcolor{darkblue}{+6.1})          & 42.4 (\textcolor{darkblue}{+5.4})          & 56.7 (\textcolor{darkblue}{+0.8})          \\
\multicolumn{2}{@{}l}{+ prompting to 2022  }    & 27.4 (\textcolor{darkblue}{+10.2})          & 40.3 (\textcolor{darkblue}{+6.2})          & 23.9 (\textcolor{darkblue}{+4.1})          & 40.5 (\textcolor{darkblue}{+3.5})          & 54.1 (\textcolor{darkred}{-1.8})           \\
\multicolumn{2}{@{}l}{+ finetuning to 2021 }      & 20.8 (\textcolor{darkblue}{+3.6})           & 39.3 (\textcolor{darkblue}{+5.2})          & \textbf{29.2 (\textcolor{darkblue}{+9.4})} & \textbf{42.5 (\textcolor{darkblue}{+5.5})} & 56.7 (\textcolor{darkblue}{+0.8})          \\
\multicolumn{2}{@{}l}{+ finetuning to 2022 }      & \textbf{27.9 (\textcolor{darkblue}{+10.7})} & \textbf{40.7 (\textcolor{darkblue}{+6.6})} & 23.7 (\textcolor{darkblue}{+3.9})          & 40.8 (\textcolor{darkblue}{+3.8})          & 55.4 (\textcolor{darkred}{-0.5})           \\
\multicolumn{2}{@{}l}{+ adaptive finetuning}      & 25.7 (\textcolor{darkblue}{+8.5})           & 40.0 (\textcolor{darkblue}{+5.9})          & 23.5 (\textcolor{darkblue}{+3.7})          & 40.9 (\textcolor{darkblue}{+3.9})          & \textbf{57.9 (\textcolor{darkblue}{+2.0})} \\ \bottomrule
\end{tabular}
\caption{Performance of temporally unaligned and aligned model on the TAQA dataset. Here, we mainly target 2021 and 2022 for recency alignment, and we evaluate the aligned LMs based on the target-year ($\mathcal{F}^{2022}$ and $\mathcal{F}^{2021}$), temporal-decayed ($\mathcal{F}^{2022}_{\text{decay}}$ and $\mathcal{F}^{2021}_{\text{decay}}$), and historical-max ($\mathcal{F}_{\text{max}}$) F1 scores, as defined in \S\ref{subsec:metrics}. 
}
\label{tab:main-results}
\end{table*}

This section describes our experiments on aligning LLaMa2 using our proposed methods on the \ourdataabbr{} dataset. Since the pretraining cutoff of LLaMa2 is September 2022, we try aligning LLaMa2 to both 2022, which it has seen partially, and 2021, which it should have seen entirely. We first introduce several baselines (\S\ref{subsec:baselines}), present the effectiveness of aligning LLaMa2 (\S\ref{subsec:main_results}), ablate different data selection strategies (\S\ref{subsec:data-selection}), and finally conduct various analyses to verify the improvement (\S\ref{subsec:improvement_analysis}).

\subsection{Baselines}
\label{subsec:baselines}

\paragraph{Unaligned models} 
To demonstrate the performance of LMs after pretraining, we choose several representative LMs, including LLaMa1-65B~\citep{touvron2023llama}, LLaMa2-70B~\citep{touvron2023llama2}, and GPT-3~\citep{brown2020language} (the ``Davinci-002'' engine using OpenAI's API). To make them answer questions in the right format, we prompt them with five time-insensitive QA examples, presented in Appendix \S\ref{sec:prompting_details}.

\paragraph{Normally aligned models} 
To show the necessity of temporal alignment against other types of alignment in prior work, we additionally test ChatGPT (``gpt-3.5-turbo-0125'' engine), which went through an RLHF process \citep{ouyang2022training}, and T\"ulu2-70B \citep{ivison2023camels}, a LLaMa2 model that underwent supervised instruction tuning.

\subsection{Results of Recency Alignment}
\label{subsec:main_results}

We present the results of aligning LLaMa2 to the years 2021 and 2022 in \autoref{tab:main-results}, as well as the performance of baseline models. All the \textbf{temporally aligned models perform better than unaligned and normally aligned models} in both target years, suggesting the possibility and effectiveness of our proposed temporal alignment. Among all the methods, \textbf{finetuning to the target year performs better than just prompting}. Given that we do not include any time information in the prompt for finetuned models, this suggests that the model is learning to adjust its internal state to activate more of its memorized recent pretraining knowledge. The \textbf{benefit of adaptive finetuning is more on historical years}, which is reflected by its best performance on $\mathcal{F}_{\text{max}}$ metric and further confirmed by its performance on a broader range of years shown in \autoref{fig:teaser}.

Moreover, comparing the two target years, our models always perform better on 2021. This could be because LLaMa2 did not see all the information in 2022 but we also speculate the information in recent years is less exposed in the pretraining data, as is also shown in \autoref{fig:teaser}. We leave the investigation into pretraining data as future work.

\subsection{What Data is More Proper for Alignment?}
\label{subsec:data-selection}

\begin{table}[htbp]
\centering
\small
\resizebox{1.0\linewidth}{!}{
\begin{tabular}{@{}lccc@{}}
\toprule
Data selection $\downarrow$ & $\mathcal{F}^{2022}$ & $\mathcal{F}_{\text{decay}}^{2022}$ & $\mathcal{F}_{\text{max}}$ \\ \midrule
All data                    & 19.8                 & 30.9                                & 44.9                       \\ \midrule
Random                      & 19.5 (\textcolor{darkred}{-0.3})                 & 30.5 (\textcolor{darkred}{-0.4})                                & 44.3 (\textcolor{darkred}{-0.6})                       \\
Popularity                  & 19.9 (\textcolor{darkblue}{+0.1})                 & 31.3 (\textcolor{darkblue}{+0.4})                                & 45.1 (\textcolor{darkblue}{+0.2})                       \\
Model confidence            & 19.5 (\textcolor{darkred}{-0.3})                 & 30.7 (\textcolor{darkred}{-0.2})                                & 44.9 (+0.0)                       \\
Model correctness           & \textbf{20.5 (\textcolor{darkblue}{+0.7})}        & \textbf{31.8 (\textcolor{darkblue}{+0.9})}                       & \textbf{45.5 (\textcolor{darkblue}{+0.6})}              \\ \bottomrule
\end{tabular}
}
\caption{Training data selection analysis with LLaMA2-13B model for optimizing the temporal alignment.}
\label{tab:data-selection}
\end{table}

We further analyze the effectiveness of our correctness-based data selection strategy (\textbf{Model Correctness}) described in \S\ref{subsec:target-year-finetuing}. 
We compare it with full-data training, as well as four other selection strategies, including random selection (\textbf{Random}), selecting the most popular questions according to page views (\textbf{Popularity}), and selecting questions that the model is mostly confident about its answer (\textbf{Model Confidence}). All the selection methods select 5000 examples from \ourdataabbr{} training set for fair comparison.

\autoref{tab:data-selection} compares these methods, proving the superior performance of correctness-based selection. This additionally confirms that the finetuning is activating models' internal knowledge, rather than injecting new knowledge into the model. Otherwise, the popularity-based or full-data training will lead to better performance.

\begin{figure*}[t!]
    \centering
    \begin{subfigure}[t]{0.48\textwidth}
        \centering
        \includegraphics[width=0.95\linewidth, trim=0 1cm 0 1cm]{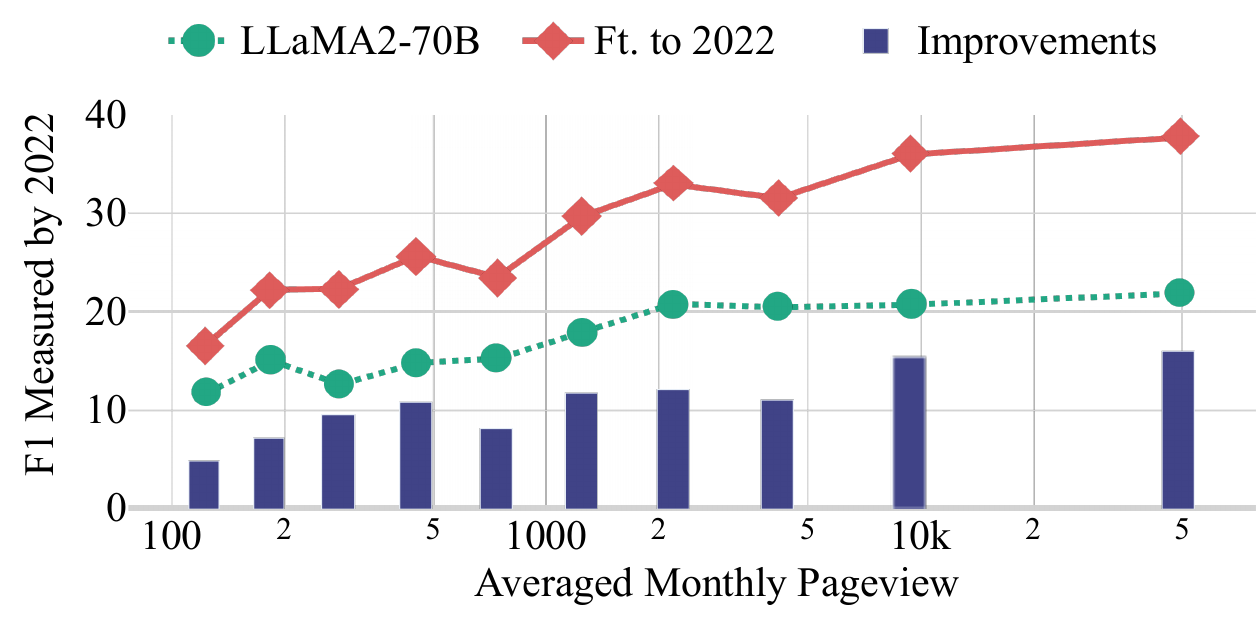}
        \label{subfig:popularity}
    \end{subfigure}%
    ~ 
    \begin{subfigure}[t]{0.48\textwidth}
        \centering
        \includegraphics[width=0.95\linewidth, trim=0 1cm 0 1cm]{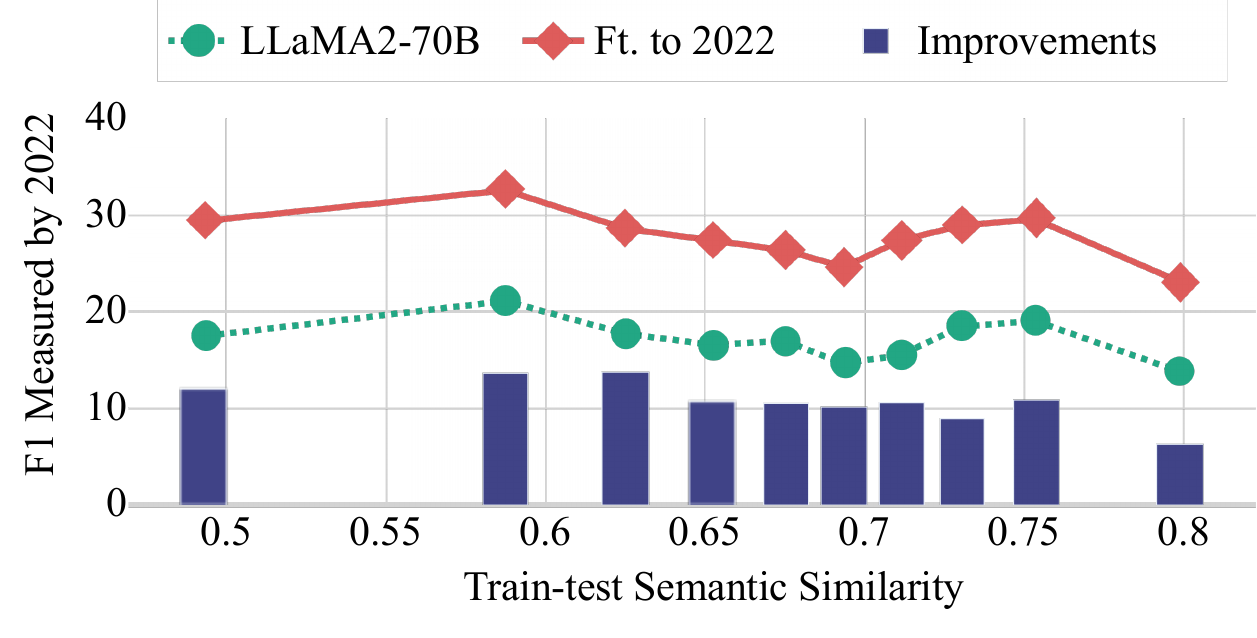}
        \label{subfig:semantic-similarity}
    \end{subfigure}
    \caption{(Left) Relationship between question popularity (measured by the pageviews of their originated Wikipedia page) and models\textquotesingle~ F1 score on them, as in 2022. (Right) Relationship between testing questions\textquotesingle~ maximum semantic similarity to the training set and models\textquotesingle~ F1 score on them, as in 2022. 
    }
    \label{fig:improvement_analysis}
\end{figure*}

\subsection{Improvement Analysis}
\label{subsec:improvement_analysis}

We provide more detailed analyses of how temporal alignment improves model predictions and confirm that the improvement is not from trivial factors.

\paragraph{Popular knowledge gets aligned better.}

We first illustrate the association between question popularity and models\textquotesingle~ improvement on them through temporal alignment, shown in \autoref{fig:improvement_analysis} (left). Overall, the target-year F1 scores of both unaligned and aligned models improve as the popularity grows, confirming that popular questions are easier for LMs to answer. Moreover, the improvement from temporal alignment also increases on popular questions, indicating that models have more potential to be aligned on popular topics, since such knowledge is better memorized during pretraining but not triggered due to the temporal chaos.

\paragraph{Improvement is not from memorizing facts in finetuning.}
The surprising effectiveness of doing temporal alignment through finetuning might be weakened if it is mainly due to some knowledge overlap between the finetuning and testing sets. To confirm this is not the reason, \autoref{fig:improvement_analysis} (right) plots the relationship between testing questions\textquotesingle semantic similarity \footnote{We use sentence-BERT embeddings~\citep{reimers-2019-sentence-bert} to compute the similarity.} to the training set and models' performance on them. We see that the improvement from alignment does not increase as the similarity grows. This implies that memorization is not the reason for better performance on \ourdataabbr.

\begin{figure}[t]
    \centering
    \includegraphics[width=\linewidth, trim=0 1cm 0 0cm]{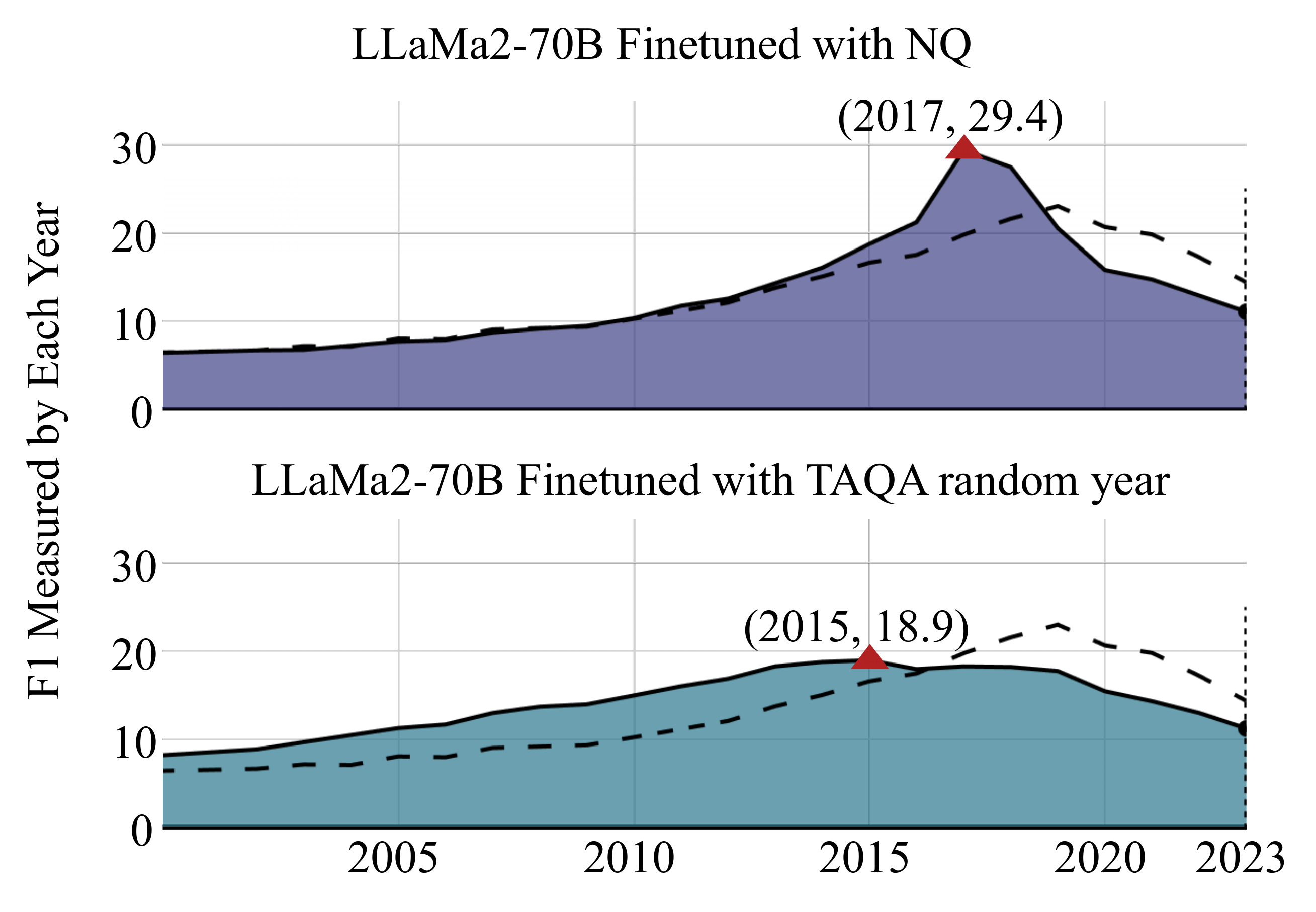}
    \caption{The temporal knowledge distribution of LMs finetuned with the NQ dataset and our {\ourdataabbr} dataset where the answers are randomly sampled. 
    }
    \label{fig:format-ablation}
\end{figure}

\paragraph{Models are aligned beyond just QA formatting.}
We further test whether the temporally aligned LMs are learning to activate recent knowledge rather than just learning to answer questions in the correct format. We conduct another two finetunings, with 1) data randomly sampled from NaturalQuestions (NQ, \citealp{kwiatkowski2019natural}) and 2) data randomly sampled from \ourdataabbr{} paired with answers picked from random years. We see from \autoref{fig:format-ablation} that both of them generally do not have the effect of aligning the model to a recent time. Interestingly, NQ actually aligns the model towards 2017, which we speculate is because the dataset was constructed using much information back then.

\subsection{Scaling Trend w.r.t. Model Size}

\begin{table}[t]
\centering
\resizebox{1.0\linewidth}{!}{
\small
\begin{tabular}{@{}llccc@{}}
\toprule
Method                     & Metric                              & 7B                                & 13B                               & 70B                                \\ \midrule
\multirow{3}{*}{Unaligned} & $\mathcal{F}^{2022}$                & 10.4                              & 13.0                              & 17.2                               \\
                           & $\mathcal{F}^{2022}_{\text{decay}}$ & 21.6                              & 25.6                              & 34.1                               \\
                           & $\mathcal{F}_{\text{max}}$          & 41.1                              & 45.5                              & 55.9                               \\ \midrule
\multirow{3}{*}{Finetuned} & $\mathcal{F}^{2022}$                & 17.6 (\textcolor{darkblue}{+7.2}) & 20.5 (\textcolor{darkblue}{+7.5}) & 27.9 (\textcolor{darkblue}{+10.7}) \\
                           & $\mathcal{F}^{2022}_{\text{decay}}$ & 28.5 (\textcolor{darkblue}{+6.9}) & 31.8 (\textcolor{darkblue}{+6.2}) & 40.7 (\textcolor{darkblue}{+6.6})  \\
                           & $\mathcal{F}_{\text{max}}$          & 43.1 (\textcolor{darkblue}{+2.0}) & 46.5 (\textcolor{darkblue}{+1.0}) & 55.4 (\textcolor{darkred}{-0.5})   \\ \midrule
\multirow{3}{*}{Adaptive}  & $\mathcal{F}^{2022}$                & 17.3 (\textcolor{darkblue}{+6.9}) & 18.9 (\textcolor{darkblue}{+5.9}) & 25.7 (\textcolor{darkblue}{+8.5})  \\
                           & $\mathcal{F}^{2022}_{\text{decay}}$ & 28.8 (\textcolor{darkblue}{+7.2}) & 32.0 (\textcolor{darkblue}{+6.4}) & 40.0 (\textcolor{darkblue}{+5.9})  \\
                           & $\mathcal{F}_{\text{max}}$          & 44.9 (\textcolor{darkblue}{+3.8}) & 50.2 (\textcolor{darkblue}{+4.7}) & 57.9 (\textcolor{darkblue}{+2.0})  \\ \bottomrule
\end{tabular}
}
\caption{Effect of LLaMA2 model sizes on temporal alignment, when finetuning to the year 2022.}
\label{tab:scaling}
\end{table}

We compare the performance of LLaMa2 70B, 13B, and 7B models before and after finetuning to analyze the effect of model size on temporal alignment. As is shown in \cref{tab:scaling}, the improvement of finetuned 7B and 13B models on target-year F1 score ($\mathcal{F}^{2022}$) are both lower than that of the 70B model.
Moreover, considering the historical-max F1 score ($\mathcal{F}^{\text{max}}$), the 70B model sees diminishing improvement after temporal alignment finetuning, while 7B and 13B models keep benefiting from finetuning. We speculate that the reason is the 70B pretrained model memorizes more factual knowledge, which might suffer from more forgetting.

\section{Further Analysis: To What Year Are LMs Most Readily Alignable?}
\label{sec:align-to-historical}

\begin{figure*}[t!]
    \centering
    \begin{subfigure}[t]{0.7\textwidth}
        \centering
        \includegraphics[width=0.9\linewidth, trim=0 0.5cm 0 1cm]{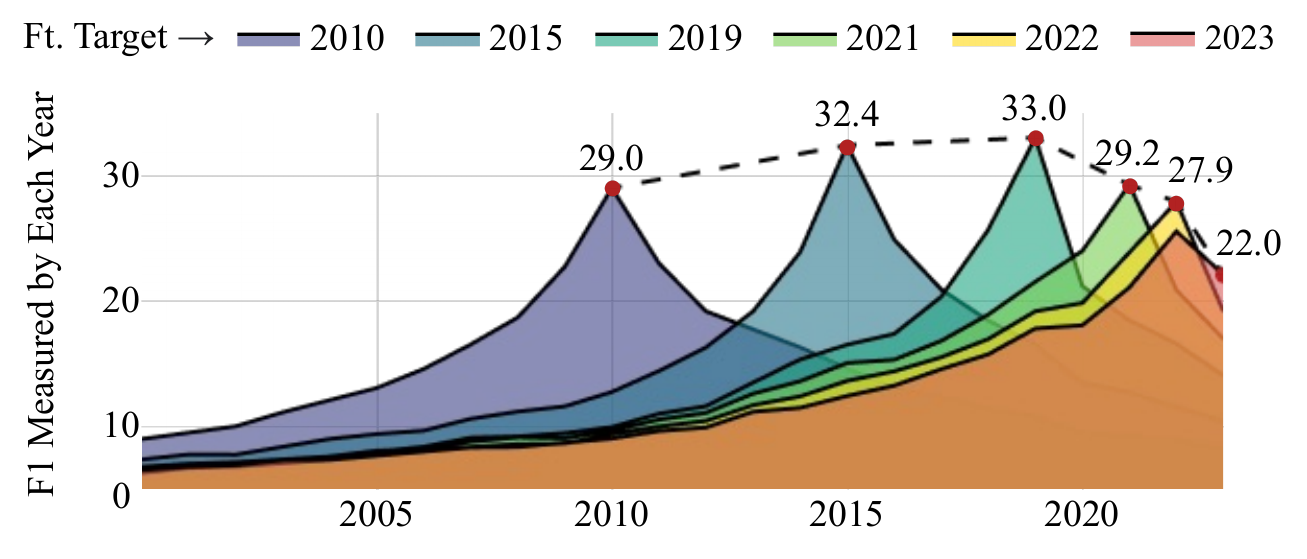}
        \label{subfig:knowledge-distribution-target-year}
    \end{subfigure}%
    ~ 
    \begin{subfigure}[t]{0.28\textwidth}
        \centering
        \includegraphics[width=0.92\linewidth, trim=0.5cm 0.5cm 0.5cm 1cm]{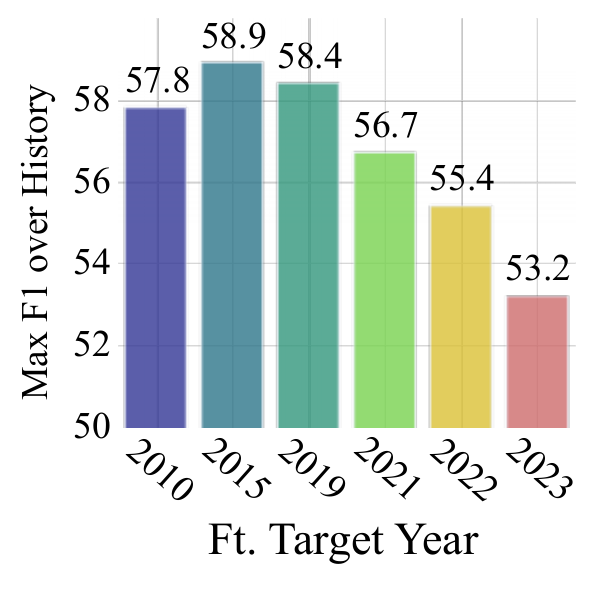}
        \label{subfig:f-max-target-year}
    \end{subfigure}
    \caption{
        LMs' F1 score in every year between 2000-2023 (left) along with their historical-max F1 scores $\mathcal{F}^{\text{max}}$ (right) after being finetuned to 2010, 2015, 2019, 2021, 2022, and 2023. The dotted line shows the trend of each aligning year's F1 score evaluated with that year's answers. 
    }
    \label{fig:target-year}
\end{figure*}

Aside from aligning LMs to a recent time, we are also curious about which year, historically speaking, an LM is most readily alignable to and achieve the best performance. We investigate this by finetuning LLaMa-70B to different years.
\cref{fig:target-year} shows that 2019 is the best year for aligning the LLaMa2-70B model instead of the year of the pretraining cutoff. Meanwhile, the performance of models finetuned with data whose answers are true in 2010, 2015, and 2019 are all better than in recent years (2022 and 2023). We believe this is because the information about the most recent factual knowledge is too short in existing pretraining corpora, so it is more challenging to align the internal knowledge of LMs to recent times than slightly further to the past. 

Furthermore, if considering the historical maximum $\mathcal{F}_{\text{max}}$, the target year of 2015 performs the best among all our target year choices. We argue that the reason could be factual knowledge in 2015 aligns the most with LLaMa2-70B's internal (but unactivated) knowledge learned from pretraining, and therefore, finetuning the LM on {\ourdataabbr} where the answers were true in 2015 causes the least contradiction and thus makes the LM the least likely to produce historically-incorrect answers.

\section{Related Work}
\label{sec:related}

\paragraph{Temporal Misalignment in LMs}
Temporal misalignment is a much-studied issue for LMs. For instance, \citet{NEURIPS2021_f5bf0ba0} pointed out that LMs suffer from the temporal gap between their training and deployment data, and \citet{jang-etal-2022-temporalwiki} proposed the benchmark of training and evaluating LMs throughout time. \citet{luu2022time} showed that temporal misalignment also hurts LMs' performance on downstream finetuning tasks, and following research indicated this effect can be task-dependent~\citep{agarwal2022temporal}. Recent research suggested that chaotic pretraining corpus is one of the reasons for LMs' temporal misalignment~\citep{longpre2023pretrainers}, and LMs can represent temporal knowledge learned from pretraining in their internal states~\citep{gurnee2023language}. These findings open up the possibility of aligning models to a specific time. Different techniques have been proposed to tackle the temporal misalignment issue~\citep{zhang2023large}, including continual pretraining~\citep{jin-etal-2022-lifelong-pretraining,loureiro-etal-2022-timelms,ke2023continual}, knowledge editing~\citep{mitchell2022fast,meng2023massediting}, and retrieval augmentation~\citep{zhang2023mitigating,vu2023freshllms}. These methods edit LMs' knowledge instead of eliciting the knowledge already learned from pretraining, as we aim to do in this work.

\paragraph{Temporal QA and Reasoning}
Temporal QA and reasoning are tasks where the correct answers change through time~\citep{Figueroa2010SurfaceLM}, e.g., ``Who is the president of the United States''.
Prior research tried using knowledge graphs to tackle these tasks~\citep{10.1145/3459637.3482416,saxena-etal-2021-question,shang-etal-2022-improving}. As modern LMs can memorize factual knowledge, new benchmarks~\citep{chen2021dataset,dhingra2022time,tan2023towards,wei2023menatqa} address the challenge of leveraging LMs to answer questions with contextual temporal mentions provided (e.g., ``in 2023'', ``after'' certain events, etc.). However, such question structure can sound unnatural, and these studies do not explicitly focus on questions whose answers have changed in recent years.
Other news article-based datasets like RealtimeQA~\citep{kasai2022realtime} and StreamingQA~\citep{liska2022streamingqa}, despite having more natural questions, still lack questions whose answers can change multiple times, rather than a single time. In contrast, our {\ourdataabbr} dataset curates natural questions without temporal information but with at least five different answers after 2000. 
Several approaches have been proposed to tackle the temporal QA and reasoning challenges~\citep{son-oh-2023-time,tan2023towards}, yet these approaches require dedicated training for enhancing LMs' understanding of temporal mentions, which are missing in most natural questions.

\section{Conclusion}
\label{sec:conclusion}

We build \ourdataabbr{}, a QA dataset with time-sensitive questions during a recent time period (2020--2023). Using this dataset, we quantitatively show that LMs can respond to questions chaotically in terms of the question's placement in time.
We thus propose the concept of temporal alignment and a set of methods for it. Extensive experiments demonstrate the possibility and effectiveness of aligning model internal knowledge temporally, both to a recent time and to a historical time.

\section{Limitations}

Throughout the development of this project, we identified several limitations that invite additional investigation. Firstly, in our dataset construction, we use GPT-4 to general natural language questions from tabular Wikipedia data without human annotation. While we do adopt quality control policies, it may still be the case that there are noisy generations. These noisy generations, where the question's text content might not be consistent with the quality of a human-written question, may impact the usefulness of such data in the alignment process and thus the effectiveness overall.
Secondly, it is difficult to determine the exact date and time that knowledge changes. This, coupled with the fact the exact pretraining corpora of LLaMa2 and similar models are unknown, means we cannot determine whether an LM has observed certain knowledge. This can introduce inaccuracies to scoring when measuring both the original year the model is internally aligned to, and the year it is aligned to after our methods are applied. Thirdly, we limit ourselves to the task of short-form question answering in this work for easy evaluation. We anticipate temporal alignment would have more interesting effects on a broader range of tasks that are time-sensitive but we leave the exploration for future work. Finally, we note that this study only considers English and English-focused LMs; the applicability of the approaches to other languages is left for future exploration.

\section{Ethics Statement}
The authors recognize the impact of LMs as components of real-world systems with real-world effects. We wish for our proposed methods make these systems more reliable and useful regarding temporal knowledge for the casual use of LMs and LMs for the professional and research communities. While LMs are still imperfect, our methods of alignment can be used to create more robust systems and mitigate the presentation of inaccurate information to users, which can significantly influence attitudes toward these technologies and the decisions of people using them.
We recognize and uphold the ACL Ethics Policy and present the motivation, methodology, and results of our work as accurately as possible.

\section*{Acknowledgements}
We thank labmates at UWNLP for their constructive feedback and
intellectual support. This work was partially supported by the Office of Naval Research under MURI grant N00014-18-1-2670, and NSF IIS-2044660, an Allen Investigator Distinguished
award. This research is also supported by Cloud TPUs from Google's TPU Research Cloud (TRC) and usage credits from OpenAI. We also thank the Beaker team at AI2, who provided the essential computational infrastructure for our experiments.

\bibliography{anthology, acl_custom}

\appendix

\onecolumn
\begin{center}
{\Large \textbf{Supplemental Material}}
\end{center}

\section{Time-Aware Prompting Details and Analysis}
\label{sec:prompting_details}

During the inference of unaligned and normally-aligned LMs, we add five time-insensitive few-shot examples in the prompt as shown in the third row of \autoref{tab:insensitive-wotime-prompt} (denoted as ``insensitive 5-shot without time'') to ensure that the LMs would respond to questions with short answers. For time-aware prompting, we propose that we can either specify the year information explicitly in the prompt (\autoref{tab:insensitive-wtime-prompt}), or use a few time-sensitive examples in the prompt with answers from the target year (\autoref{tab:sensitive-wotime-prompt}). Our final strategy uses both types of information (\autoref{tab:sensitive-wtime-prompt}).

We analyzed the performance of different time-aware prompting strategies applied to LLaMa2-70B models with the target years 2021 and 2022 in \autoref{tab:prompt-analysis}. In general, \textbf{all prompting techniques improve the model's performance} in both target years, indicating that either time-sensitive examples or time (year) mention help LMs elicit temporal knowledge better. In addition, using time-sensitive examples plus the year-mention in the prompt achieves higher target-year and temporal-decayed performance. Meanwhile, \textbf{prompting LMs to be aligned to 2022 causes more performance degradation on $\mathcal{F}_{\text{max}}$}, suggesting that prompting LMs to be aligned to a more recent year could lead to more QA errors. 

\begin{table*}[h]
\centering
\begin{tabular}{@{}lccccc@{}}
\toprule
Prompting Strategy                      & $\mathcal{F}^{2022}$ & $\mathcal{F}_{\text{decay}}^{2022}$ & $\mathcal{F}^{2021}$ & $\mathcal{F}_{\text{decay}}^{2021}$ & $\mathcal{F}_{\text{max}}$ \\ \midrule
Insensitive 5-shot w/o time        & 17.2               & 34.1                              & 19.8               & 37.0                              & 55.9                     \\ \midrule
Insensitive 5-shot w/ time to 2022 & 24.4 (\textcolor{darkblue}{+7.2})               & 37.9 (\textcolor{darkblue}{+3.8})                              & 24.4 (\textcolor{darkblue}{+4.6})               & 39.1 (\textcolor{darkblue}{+2.1})                              & 52.8 (\textcolor{darkred}{-3.1})                     \\
Sensitive 5-shot w/o time to 2022  & \textbf{27.6 (\textcolor{darkblue}{+10.4})}      & 40.2 (\textcolor{darkblue}{+6.1})                              & 22.8 (\textcolor{darkblue}{+3.0})               & 39.9 (\textcolor{darkblue}{+2.9})                              & 53.6 (\textcolor{darkred}{-2.3})                    \\

Sensitive 5-shot w/ time to 2022   & 27.4 (\textcolor{darkblue}{+10.2})               & \textbf{40.3 (\textcolor{darkblue}{+6.2})}                     & 23.9 (\textcolor{darkblue}{+4.1})               & 40.5 (\textcolor{darkblue}{+3.5})                              & 54.1 (\textcolor{darkred}{-1.8})                     \\ \midrule
Insensitive 5-shot w/ time to 2021 & 19.2 (\textcolor{darkblue}{+2.0})               & 36.8 (\textcolor{darkblue}{+2.7})                              & 25.3 (\textcolor{darkblue}{+5.5})               & 39.9 (\textcolor{darkblue}{+2.9})                              & 53.7 (\textcolor{darkred}{-2.2})                     \\
Sensitive 5-shot w/o time to 2021  & 19.4 (\textcolor{darkblue}{+2.2})               & 38.5 (\textcolor{darkblue}{+4.4})                              & 25.0 (\textcolor{darkblue}{+5.2})               & 41.9 (\textcolor{darkblue}{+4.9})                              & \textbf{56.8 (\textcolor{darkblue}{+0.9})}            \\
Sensitive 5-shot w/ time to 2021   & 19.7 (\textcolor{darkblue}{+2.5})               & 38.9 (\textcolor{darkblue}{+4.8})                              & \textbf{25.9 (\textcolor{darkblue}{+6.1})}      & \textbf{42.4 (\textcolor{darkblue}{+5.4})}                     & 56.7 (\textcolor{darkblue}{+0.8})            \\
\bottomrule
\end{tabular}
\caption{Performance of different time-aware prompting strategies applied to LLaMa2-70B models on the TAQA test set. Same as \S\ref{subsec:main_results}, we evaluate the LMs based on the target-year ($\mathcal{F}^{2022}$ and $\mathcal{F}^{2021}$), temporal-decayed ($\mathcal{F}^{2022}_{\text{decay}}$ and $\mathcal{F}^{2021}_{\text{decay}}$), and historical-max ($\mathcal{F}_{\text{max}}$) F1 scores, as defined in \S\ref{subsec:metrics}. }
\label{tab:prompt-analysis}
\end{table*}

\begin{table*}[ht]
    \centering
    \small
    \noindent\fbox{%
    \begin{minipage}{\dimexpr\linewidth\fboxsep-2\fboxrule} 
\tt 
\textbf{Prompt}: \newline
Answer the following question: What is the capital of France?\newline The answer is: Paris\newline \newline Answer the following question: Who wrote Harry Potter?\newline The answer is: J.K. Rowling\newline \newline Answer the following question: Where did the Titanic sink?\newline The answer is: Atlantic Ocean \newline \newline Answer the following question: What is the gravity of earth?\newline The answer is: 9.807 m/s\textasciicircum2 \newline \newline Answer the following question: Is the speed of light faster than the speed of sound?\newline The answer is: Yes \newline \newline Answer the following question: Which team won the UEFA Europa League?\newline The answer is:
\newline\newline
\textbf{Expected Output}: \newline
Eintracht Frankfurt
    \end{minipage}
}

\caption{Our prompt for prompting the pretrained LM using the insensitive w/o time strategy for time-sensitive QA tasks.}
\label{tab:insensitive-wotime-prompt}
\end{table*}

\begin{table*}[ht]
    \centering
    \small
    \noindent\fbox{%
    \begin{minipage}{\dimexpr\linewidth\fboxsep-2\fboxrule} 
\tt 
\textbf{Prompt}: \newline
Answer the following question: What is the capital of France?\newline As of year 2022, the answer is: Paris\newline \newline Answer the following question: Who wrote Harry Potter?\newline As of year 2022, the answer is: J.K. Rowling\newline \newline Answer the following question: Where did the Titanic sink?\newline As of year 2022, the answer is: Atlantic Ocean \newline \newline Answer the following question: What is the gravity of earth?\newline As of year 2022, the answer is: 9.807 m/s\textasciicircum2 \newline \newline Answer the following question: Is the speed of light faster than the speed of sound?\newline As of year 2022, the answer is: Yes \newline \newline Answer the following question: Which team won the UEFA Europa League?\newline As of year 2022, the answer is:
\newline\newline
\textbf{Expected Output}: \newline
Eintracht Frankfurt
    \end{minipage}
}

\caption{Our prompt for prompting the pretrained LM using the insensitive w/ time strategy for time-sensitive QA tasks.}
\label{tab:insensitive-wtime-prompt}
\end{table*}

\begin{table*}[ht]
    \centering
    \small
    \noindent\fbox{%
    \begin{minipage}{\dimexpr\linewidth\fboxsep-2\fboxrule} 
\tt 
\textbf{Prompt}: \newline
Answer the following question: Which Hindi film has the highest domestic net collection currently?\newline The answer is: Brahmāstra: Part One – Shiva\newline \newline Answer the following question: Where is the NHL Winter Classic taking place?\newline The answer is: Target Field \newline \newline Answer the following question: Who are the current drivers for the Mercedes-Benz Formula One team?\newline The answer is: Lewis Hamilton George Russell \newline \newline Answer the following question: Who received the Player of the Game award for offense in the most recent Rose Bowl Game?\newline The answer is: Jaxon Smith-Njigba \newline \newline Answer the following question: Where was the final of the last FIFA Club World Cup held?\newline The answer is: Prince Moulay Abdellah Stadium, Rabat \newline \newline Answer the following question: Which team won the UEFA Europa League?\newline The answer is:
\newline\newline
\textbf{Expected Output}: \newline
Eintracht Frankfurt
    \end{minipage}
}

\caption{Our prompt for prompting the pretrained LM using the sensitive w/o time strategy for time-sensitive QA tasks.}
\label{tab:sensitive-wotime-prompt}
\end{table*}

\begin{table*}[ht]
    \centering
    \small
    \noindent\fbox{%
    \begin{minipage}{\dimexpr\linewidth\fboxsep-2\fboxrule} 
\tt 
\textbf{Prompt}: \newline
Answer the following question: Which Hindi film has the highest domestic net collection currently?\newline As of year 2022, the answer is: Brahmāstra: Part One – Shiva\newline \newline Answer the following question: Where is the NHL Winter Classic taking place?\newline As of year 2022, the answer is: Target Field \newline \newline Answer the following question: Who are the current drivers for the Mercedes-Benz Formula One team?\newline As of year 2022, the answer is: Lewis Hamilton George Russell \newline \newline Answer the following question: Who received the Player of the Game award for offense in the most recent Rose Bowl Game?\newline As of year 2022, the answer is: Jaxon Smith-Njigba \newline \newline Answer the following question: Where was the final of the last FIFA Club World Cup held?\newline As of year 2022, the answer is: Prince Moulay Abdellah Stadium, Rabat \newline \newline Answer the following question: Which team won the UEFA Europa League?\newline As of year 2022, the answer is:
\newline\newline
\textbf{Expected Output}: \newline
Eintracht Frankfurt
    \end{minipage}
}

\caption{Our prompt for prompting the pretrained LM using the sensitive w/ time strategy for time-sensitive QA tasks.}
\label{tab:sensitive-wtime-prompt}
\end{table*}

\clearpage

\section{Details of Finetuning Methods}
\label{sec:training_details}

\subsection{Formatting of Training Examples}
When training and inferring the target-year finetuning models, as shown in \autoref{tab:standard-qa-prompt}, we do not use specific instructions or prompts but query them with the questions directly. For temporal-adaptive finetuning setup, as we want the aligned LM to decide which year's knowledge to use for answering each question, we use the prompt shown in \autoref{tab:temporal-adaptive-prompt}, as described in \S\ref{subsec:temporal-adaptive}. Moreover, when calculating the loss during training, we consider the answer only for the target-year finetuning model; for temporal-adaptive finetuning models, we calculate the loss on the whole generated sequence.

\begin{table*}[ht]
    \centering
    \small
    \noindent\fbox{%
    \begin{minipage}{\dimexpr\linewidth-50\fboxsep-2\fboxrule} 
\tt 
\textbf{Prompt}: \newline
Answer the following question: Which team won the UEFA Europa League? \newline The answer is:\newline
\newline
\textbf{Expected Output}: \newline
Eintracht Frankfurt
    \end{minipage}
}

\caption{Our prompt and the expected outputs for finetuning LMs on our TAQA dataset. We use the same prompts during the inference of the finetuning-aligned models.}
\label{tab:standard-qa-prompt}
\end{table*}

\begin{table*}[ht]
    \centering
    \small
    \noindent\fbox{%
    \begin{minipage}{\dimexpr\linewidth-50\fboxsep-2\fboxrule} 
\tt 
\textbf{Prompt}: \newline Answer the following question: Which team won the UEFA Europa League? \newline \newline \textbf{Expected Output}: \newline Based on my latest knowledge for this question from year \textbf{2022}, the answer is: \textbf{Eintracht Frankfurt}
    \end{minipage}
}

\caption{Our prompt and the expected outputs for finetuning LMs with our temporal-adaptive strategy on the TAQA dataset. We use the same prompts during the inference of the finetuning-aligned models.}
\label{tab:temporal-adaptive-prompt}
\end{table*}

\subsection{Training Details}

All 70B model finetuning was done on a 256-chip TPU v3 pod, with a codebase based off EasyLM~\citep{geng2023easylm}. The finetuning of smaller models and all evaluations were done on an internal A100 80GB cluster. 
At the start of our finetuning, we did a hyperparameter search using LLaMa2 13B model on the development set of \ourdataabbr. We then keep the same set of hyperparameters for all our finetuning experiments.
These hyperparameters are shown in \autoref{tab:hyperparameters}.
\begin{table}[h]
    \centering
    \begin{tabular}{cc}
    \toprule
    Hyperparameter & Value \\
    \midrule
      Precision   &  BFloat16 \\
      Epochs   &  2 \\
      Learning rate  & 5e-6 \\
      Warmup ratio   & 0.03 \\
      Weigtht decay   & Linear \\
      Max. seq. length  & 128 \\
      Batch size       & 128 \\
    \bottomrule
    \end{tabular}
    \caption{Finetuing hyperparameters}
    \label{tab:hyperparameters}
\end{table}

\subsection{Evaluation}
When evaluating temporal-adaptive aligned LMs, since the models always generate the prefix demonstrating the temporal information before answering the question, we extract the answers from the model outputs after ``the answer is:'' as the model predictions. As target-year finetuning models directly answer time-sensitive questions, we treat the model's raw outputs as predicted answers.

\clearpage

\section{Question Generation Details} 
\label{sec:qg_details}
We use eight few-shot examples to prompt the GPT-4-turbo model (the ``gpt-4-1106-preview'' engine) for question generation as it is updated with knowledge in 2023, with relatively shorter latency and lower cost. We demonstrate our prompt in \autoref{tab:qg_prompt}, \autoref{tab:qg_prompt_2}, \autoref{tab:qg_prompt_3}, and \autoref{tab:qg_prompt_4}. We set the temperature to 1.0 for all our queries, hoping to gather more diverse questions in styles and formats. We specifically curated the few-shot examples in vast domains, including sports, films, law, politics, etc. in order to ensure GPT-4 can learn the correct task format in most scenarios when being queried with contents from various Wikipedia pages. We also deliberately add one example (Example 8) without any valid question to be generated to make sure GPT-4 will reject to generate time-insensitive questions when being queried with such unsuitable tables.

\begin{table*}[ht]
    \centering
    \small
    \noindent\fbox{%
    \begin{minipage}{\dimexpr\linewidth\fboxsep-2\fboxrule} 
\tt 
Below is a table in CSV format separated with commas ",". Based on this table's information along with its description and abstracts, please raise up to ten questions that the answers are different in 2010, 2019, and 2023. After raising the question, please get answers for those time points from the table. Please make sure the answers should be totally different without overlapping. Please only raise questions regarding the latest status of those time points instead of the history, while also do not explicitly mention the time information in the question to make them more natural. Do not ask questions cannot be answered based on the information provided in the table. Make sure the full and explicit names of related entities are used in the question based on the description in the prompt and do not use pronouns. Each question should contain only one column's information and no other columns' values shall be mentioned. Please select columns and generation questions based on the Query data, but not the examples. Do not answer the questions in your response. Please reply with the same format as the examples. Do not use a column to generate questions more than once. Do not use other columns' information to be the condition or the clause of the question.\newline Example 1:\newline Table description: this table is about List of highest-grossing films Timeline of highest-grossing films.\newline Table content:\newline Established,Title,Record-setting gross\newline 1998,Titanic,"\$1,843,373,318"\newline 2010,Avatar,"\$2,743,577,587"\newline 2010,Avatar,"\$2,788,416,135"\newline 2019,Avengers: Endgame,"\$2,797,501,328"\newline 2022,Avatar,"\$2,923,706,026"\newline \newline Generated questions for Example 1 asking for information in a specific column:\newline Column 0: Title\newline Question 0: What is the highest grossing film of all time?\newline \newline Column 1: Record-setting gross\newline Question 1: What is the record-setting gross of the highest grossing film of all time?\newline \newline \newline Example 2:\newline Table description: this table is about List of NBA champions Champions.\newline Table content:\newline Year,Western champion,Result,Eastern champion,Finals MVP,Coach (1),Coach (2)\newline 2000,"Los Angeles Lakers (1) (25, 12–13)",4–2,"Indiana Pacers (1) (1, 0–1)",Shaquille O'Neal,Phil Jackson,Larry Bird\newline 2010,"Los Angeles Lakers (1) (31, 16–15)",4–3,"Boston Celtics (4) (21, 17–4)",Kobe Bryant,Phil Jackson,Doc Rivers\newline 2019,"Golden State Warriors (1) (11, 6–5)",2–4,"Toronto Raptors (2) (1, 1–0)",Kawhi Leonard,Steve Kerr,Nick Nurse\newline 2023,"Denver Nuggets (1) (1, 1–0)",4–1,"Miami Heat (8) (7, 3–4)",Nikola Jokić,Michael Malone,Erik Spoelstra\newline \newline Generated questions for Example 2 asking for information in a specific column:\newline Column 0: Eastern champion\newline Question 0: Which team was the eastern champion played in the last NBA final?\newline \newline Column 1: Finals MVP\newline Question 1: Who's the MVP of the last NBA final?\newline \newline \newline Example 3:\newline Table description: this table is about Chris Pratt Filmography, Film.\newline Table content:\newline Year,Title,Role,Notes\newline 2000,Cursed Part 3,Devon,Short film\newline 2009,Bride Wars,Fletcher Flemson,\newline 2009,Deep in the Valley,Lester Watts,\newline 2009,Jennifer's Body,Roman Duda,\newline 2009,The Multi-Hyphenate,Chris,Short film\newline 2019,The Lego Movie 2: The Second Part,"Emmet Brickowski, Rex Dangervest",Voice\newline 2019,The Kid,Grant Cutler,\newline 2019,Avengers: Endgame,Peter Quill / Star-Lord,\newline 2023,The Super Mario Bros. Movie,Mario,Voice\newline 2023,Guardians of the Galaxy Vol. 3,Peter Quill / Star-Lord,\newline \newline
    \end{minipage}
}

\caption{Sampled prompt for question generation with GPT-4 (part 1).}
\label{tab:qg_prompt}
\end{table*}

\begin{table*}[ht]
    \centering
    \small
    \noindent\fbox{%
    \begin{minipage}{\dimexpr\linewidth\fboxsep-2\fboxrule} 
\tt 
Generated questions for Example 3 asking for information in a specific column:\newline Column 0: Title\newline Question 0: What's the movie that Chris Pratt most recently starred in?\newline \newline Column 1: Role\newline Question 1: What's the role that Chris Pratt played in the last movie?\newline \newline \newline Example 4:\newline Table description: this table is about Judicial Committee of the Privy Council Jurisdiction, Jurisdiction removed.\newline Table content:\newline Country,Date,Abolishing statute,New court of final appeal,Notes\newline The Gambia,1998,1997 Constitution of the Gambia,Supreme Court,"A restructure of the Gambian judiciary by Yahya Jammeh, which made the Supreme Court of The Gambia the highest court instead of being below the Court of Appeal of the Gambia as was the case under the 1970 Constitution of the Gambia."\newline BLZ,2010,"Belize Constitution (Seventh Amendment) Act, 2010",Caribbean Court of Justice,\newline DMA,2015,"Constitution of Dominica (Amendment) Act, 2014",Caribbean Court of Justice,\newline LCA,2023,"Constitution of Saint Lucia (Amendment) Act, 2023",Caribbean Court of Justice,\newline \newline Generated questions for Example 4 asking for information in a specific column:\newline Column 0: Country\newline Question 0: What country abolished the Judicial Committee of the Privy Council of the Uinted Kingdom most recently?\newline \newline Column 1: Abolishing statute\newline Question 1: What is the abolishing statute of the Judicial Committee of the Privy Council of the United Kingdom in the last country that abolished it?\newline \newline \newline Example 5:\newline Table description: this table is about National People's Congress Membership, Membership of previous National People's Congresses.\newline Table content:\newline Congress,Year,Total deputies,Female deputies,Female \%,Minority deputies,Minority \%\newline Ninth,1998,2979,650,21.8,428,14.4\newline Eleventh,2008,2987,637,21.3,411,13.8\newline Thirteenth,2018,2980,742,24.9,438,14.7
\newline Fourteenth,2023,2977,790,26.5,442,14.8\newline \newline Generated questions for Example 5 asking for information in a specific column:\newline Column 0: Total deputies\newline Question 0: What is the total number of deputies in the National People's Congress of China?\newline \newline Column 1: Female deputies\newline Question 1: How many female deputies are there in China's National People's Congress?\newline \newline Column 2: Minority \%\newline Question 2: What is the ratio of deputy members from minority ethnic groups in China's National People's Congress?\newline \newline 
\newline Example 6:\newline Table description: this table is about List of justices of the South Carolina Supreme Court .\newline Table content:\newline Justice,Began active service,Ended active service,Notes\newline Costa M. Pleicones,2000,2017,Became chief justice in 2016\newline Kaye Gorenflo Hearn,2009,2022,-\newline George C. James,2017,Incumbent,-\newline D. Garrison Hill,2023,Incumbent,-\newline \newline Generated questions for Example 6 asking for information in a specific column:\newline Column 0: Justice\newline Question 0: Who recently became the chief justice of the South Carolina Supreme Court?
    \end{minipage}
}
\caption{Sampled prompt for question generation with GPT-4 (part 2).}
\label{tab:qg_prompt_2}
\end{table*}

\begin{table*}[ht]
    \centering
    \small
    \noindent\fbox{%
    \begin{minipage}{\dimexpr\linewidth-\fboxsep-2\fboxrule} 
\tt 
Example 7:\newline Table description: this table is about Zlatan Ibrahimović Career statistics, Club.
\newline Table content:
\newline Club,Season,League (Division),League (Apps),League (Goals),National cup (Apps),National cup (Goals),League cup (Apps),League cup (Goals),Continental (Apps),Continental (Goals),Other (Apps),Other (Goals),Total (Apps),Total (Goals),date\_year,start\_time,end\_time
\newline Malmö FF,2000,Superettan,26,12,3,2,—,—,—,—,—,—,29,14,2000.0,2000.0,2000.0
\newline Barcelona,2009–10,La Liga,29,16,2,1,—,—,10,4,4,0,45,21,2009.0,2009.0,2010.0\newline AC Milan (loan),2010–11,Serie A,29,14,4,3,—,—,8,4,—,—,41,21,2010.0,2010.0,2011.0\newline LA Galaxy,2019,Major League Soccer,29,30,0,0,2,1,—,—,—,—,31,31,2019.0,2019.0,2019.0\newline AC Milan,2019–20,Serie A,18,10,2,1,—,—,—,—,—,—,20,11,2019.0,2019.0,2020.0\newline AC Milan,2021–22,Serie A,23,8,0,0,—,—,4,0,—,—,27,8,2021.0,2021.0,2022.0\newline \newline Generated questions for Example 7 asking for information in a specific column:\newline Column 0: Club\newline Question 0: Which team does Zlatan Ibrahimović play for?\newline 
\newline Column 1: League (Division)\newline Question 1: Which league and division does Zlatan Ibrahimović play in?\newline \newline Column 2: League (Goals)\newline Question 2: How many goals does Zlatan Ibrahimović score in league this season?\newline \newline Column 3: League (Apps)\newline Question 3: How many games does Zlatan Ibrahimović play in league this season?\newline \newline Column 4: Total (Goals)\newline Question 4: How many goals does Zlatan Ibrahimović score in total this season?\newline \newline \newline Example 8:\newline Table description: this table is about List of Denmark women's international footballers List of players.\newline Table content:
\newline \#,Player,Pos.,Caps,Goals,Debut,Last cap,Medals\newline 94,Signe Højen Andersen,MF,25,0,2000,2004,
\newline 23,Heidi Johansen,GK,80,0,2000,2012,\newline 31,Tine Cederkvist,GK,68,0,2000,2011,
\newline 24,Gitte Andersen Gitte Andersen (footballer),DF,79,1,2000,2007,\newline 9,Cathrine Paaske Sørensen,MF,121,36,2000,2010,\newline 39,Janne Madsen,MF,61,4,2000,2009,\newline 17,Janni Arnth,DF,93,2,2010,2019,2017 EC\newline 54,Sofie Svava,MF,43,3,2019,2023,\newline 78,Emma Snerle,MF,30,2,2019,2023,
\newline 83,Mille Gejl,FW,28,7,2019,2023,\newline 78,Rikke Marie Madsen,FW,30,1,2019,2023,\newline 68,Kathrine Kühl Kathrine Møller Kühl,MF,33,1,2021,2023,\newline \newline Generated questions for Example 8 asking for information in a specific column:\newline No questions can be generated.\newline \newline Query:\newline Table description: this table is about List of European Cup and UEFA Champions League winning managers European Cup and UEFA Champions League winning managers, By year.\newline Table caption: European Cup and UEFA Champions League winning managers*\newline Table content:\newline Final,Nationality,Winning manager,Nation,Club\newline 2010,POR,José Mourinho,ITA,Inter Milan\newline 2019,GER,Jürgen Klopp,ENG,Liverpool\newline 2023,ESP,Pep Guardiola,ENG,Manchester City\newline \newline Generated questions for Query asking for information in a specific column:
    \end{minipage}
}

\caption{Sampled prompt for question generation with GPT-4 (part 3).}
\label{tab:qg_prompt_3}
\end{table*}

\begin{table*}[ht]
    \centering
    \small
    \noindent\fbox{%
    \begin{minipage}{\dimexpr\linewidth\fboxsep-2\fboxrule} 
\tt 
Column 0: Final\newline Question 0: In which UEFA Champions League final did the winning manager lead his team to victory?\newline \newline Column 1: Nationality\newline Question 1: What is the nationality of the manager who won the UEFA Champions League?\newline \newline Column 2: Winning manager\newline Question 2: Who is the winning manager of the UEFA Champions League?\newline \newline Column 3: Nation\newline Question 3: From which nation is the winning club of the UEFA Champions League?\newline \newline Column 4: Club\newline Question 4: Which club won the UEFA Champions League?
    \end{minipage}
}

\caption{Sampled response from GPT-4 for question generation.}
\label{tab:qg_prompt_4}
\end{table*}

\section{QA Post-Processing Details}
\label{sec:qa_post_process}
After extracting the answers from the Wikipedia tables based on columns selected by GPT-4, we first use a set of heuristic rules to discard noisy QA instances. Specifically, we remove the data that 1) has more than five correct answers per year on average and 2) the averaged answer length is bigger than ten. Therefore, we can ensure that the answers in our \ourdataabbr{} dataset are concise and clear, and short in format.

During QA post-processing, we first use Pyserini~\citep{Lin_etal_SIGIR2021_Pyserini} with Lucene engine to calculate the BM25 scores of each question and answer pairs in the curated dataset. 
However, since the range of BM25 scores is unbounded, we cannot directly use them to identify duplicate questions with a static threshold. As a result, we use the score normalized with the score between the question/answer to itself to stand for similarity: $\textrm{Sim}(q_i, q_j) = \textrm{BM25}(q_i, q_j) / \textrm{BM25}(q_i, q_i) \in [0, 1]$. A pair of QA instances are marked as duplicates if they fit the following criteria: 1)either question or answer similarity is greater than 0.9, or 2)their question similarity is greater than 0.8 \textbf{AND} their answer similarity is greater than 0.5. We set the first criteria's threshold relatively higher because QA instances might have almost the same questions but correspond to totally different answers, e.g., ``Who is the President of the United States?'' v.s. ``Who is the Vice President of the United States''.

After de-duplicating the QA data based on the policies above and selecting the instances based on popularity and answer occurrence, as described in \S\ref{subsec:data-selection}, we further clean the answer texts to confirm there is little noise for LM training and evaluation. In detail, we first discard QA data with answers demonstrating unavailable information, e.g., ``N/A'', ``TBA'', ``TBD'', etc. Afterward, we further clean the redundant HTML tags and Wikitext templates in the answer texts. Finally, as nationalities often appear near peoples' names on Wikipedia tables, we use country codes to identify and clean them as those columns' are designed for questions regarding people but not countries.

\clearpage

\section{Case Studies}
\label{sec:appendix}
\label{sec:appendix:case-studies}

We perform a small series of case studies to demonstrate the effectiveness of our proposed alignment techniques in practice. To make results relevant to real-world applications and create a challenging setup for our methods, we use a 6,619-question test set subset where the questions' answers changed recently. More specifically, we only use questions from our test set with ground truth answers from 2020 to 2022 that share no common element to the answers in 2019, which is the knowledge peak of the LLaMa2-70B model as shown in \autoref{fig:teaser}.

In detail, we compare the model predictions between the unaligned, target-year (2022) finetuning, and temporal-adaptive finetuning models. To simplify our analysis, we categorize the model predictions into three: 1) correct, where the model's prediction hits one of the ground truth answers in 2022; 2) misaligned, where the model's prediction hits one of the answers that are historically correct, but not in 2022; 3) incorrect, where the model's prediction does not hit any of the answers in the question's historical answer set. We present the number of answers predicted by the unaligned and target-year finetuning models categorized into the three classes above in \autoref{tab:target-year-error-analysis}. 

\begin{table}[h]
\centering
\resizebox{0.6\linewidth}{!}{
\begin{tabular}{@{}ll|rrr@{}}
\toprule
    &            & \multicolumn{3}{c}{Unaligned}    \\
    &            & Correct & Misaligned & Incorrect \\ \midrule
\multirow{3}{*}{\makecell{Target-Year \\ Finetuned}} & Correct    & 186     & 522        & 307       \\
    & Misaligned & 57      & 924        & 553       \\
    & Incorrect  & 89      & 765        & 3216      \\ \midrule
\multicolumn{2}{c|}{Total}                                                                      & 332     & 2211       & 4076      \\ \bottomrule
\end{tabular}
}
\caption{The quantity of correct, misaligned, and incorrect answers predicted by unaligned and target-year (2022) finetuned models.}
\label{tab:target-year-error-analysis}
\end{table}

\subsection{What questions can the LM answer correctly without alignment, even if their answers changed recently?}

From \autoref{tab:target-year-error-analysis}, we find that the unaligned LLaMa2-70B model can still answer 332 time-sensitive questions correctly. We show four examples from this subset in \autoref{tab:unaligned-correct}. 

\begin{table*}[ht]
\renewcommand{\arraystretch}{1.4}
\renewcommand{\cellalign}{lc}
\centering
\small
\resizebox{1.0\linewidth}{!}{
\begin{tabular}{@{}lll@{}}
\toprule
Question                                                                                                                                              & Unaligned Model's Answer                         & Answer Correct Years \\ \midrule
\makecell{Who is the current captain of Leicester City F.C.?}   & Kasper Schmeichel    & 2021, 2022   \\ \midrule
\makecell{What is the official United Nations theme for\\ International Women's Day?}                   & Gender equality today for a sustainable tomorrow & 2022                 \\ \midrule
\makecell{What character does Jeri Ryan portray \\in her current television series?}                   & Seven of Nine                        & 2020, 2021, 2022                 \\ \midrule
\makecell{Who holds the position of offensive coordinator\\ for the Pittsburgh Steelers at the moment?} & Matt Canada                                      & 2021, 2022, 2023     \\ 
\bottomrule
\end{tabular}
}
\caption{Examples of questions that can be answered correctly as of 2022 by the unaligned LLaMa2-70B model, even if their answers differ from those in 2019.}
\label{tab:unaligned-correct}
\end{table*}

\begin{figure}[htbp]
    \centering
    \includegraphics[width=0.85\linewidth]{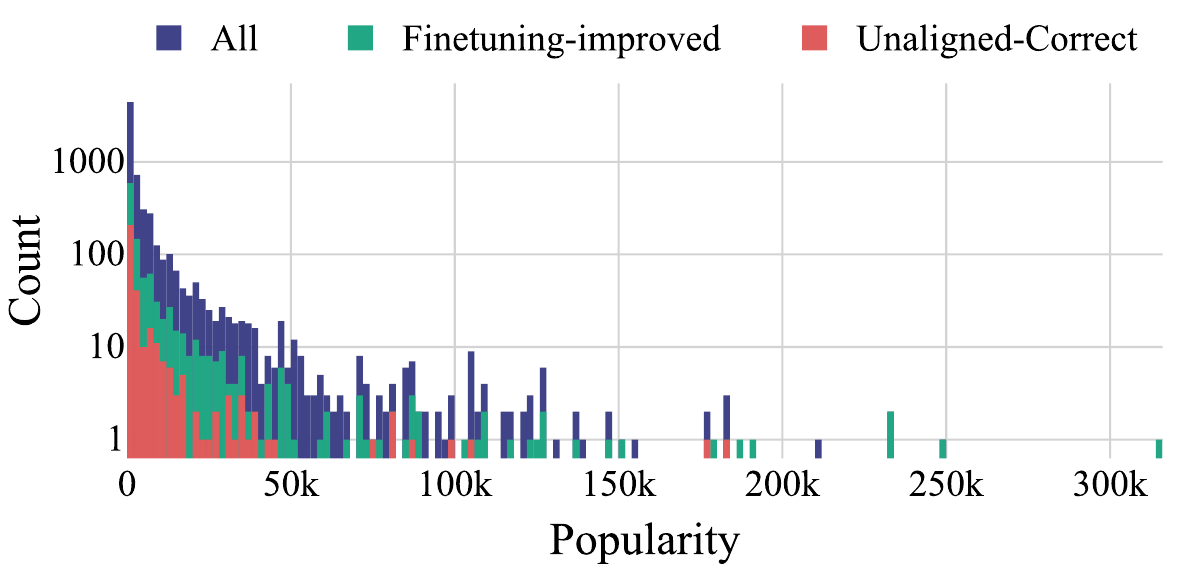}
    \caption{The popularity distribution of the questions with answers changed after 2019, and among those who can be answered correctly by the unaligned LLaMa2-70B model, and whose answers improved because of target-year finetuning.}
    \label{fig:unaligned-correct-popularity}
\end{figure}

In addition, we want to check what kinds of questions are easier to answer correctly by the unaligned LM, even if their answers are different from 2019. From \autoref{fig:unaligned-correct-popularity}, we find that there is no obvious difference in the popularity distribution between all questions whose answers changed after 2019 and among those that can be answered correctly by the unaligned model. Hence, we claim that popularity is not an important factor in determining whether a time-sensitive question can be answered correctly by an unaligned LM. Moreover, we group the questions whose answers changed after 2019 by their answer-changing years as shown in \autoref{tab:earliest-correct-year}. We notice that there are 25.3\% and 27.7\% of the questions whose answers became true in 2020 and 2021, respectively, among those questions can be answered correctly by the unaligned models, while the ratios of those among the whole subset are 9.2\% and 23.3\%. The result implies that time-sensitive questions with answers changed earlier can be easier for unaligned models. Furthermore, we hypothesize that a more profound reason is that those QA-pairs, which started to become true earlier, have more passages in LMs' pretraining corpora to describe them, so LMs can answer them correctly after observing the knowledge multiple times during pretraining.

\begin{table*}[htbp]
\centering
\begin{tabular}{@{}lccc@{}}
\toprule
                          & \multicolumn{3}{c}{Earliest Correct Year}                                                               \\
                          & 2020                            & 2021                              & 2022                              \\ \midrule
Unaligned Correct         & 84 (25.3\%)                     & 125 (37.7\%)                      & 123 (37.0\%)                      \\
Target-year Correct       & 152 (15.0\%)                    & 332 (32.7\%)                      & 531 (52.3\%)                      \\
Temporal-adaptive Correct & 136 (15.3\%)                    & 307 (34.7\%)                      & 443 (50.5\%)                      \\
Total                     & \multicolumn{1}{l}{610 (9.2\%)} & \multicolumn{1}{l}{1540 (23.3\%)} & \multicolumn{1}{l}{4469 (67.5\%)} \\ \bottomrule
\end{tabular}
\caption{The number of questions whose answers changed after 2019, grouped by the earliest year that their answers in 2022 became true.}
\label{tab:earliest-correct-year}
\end{table*}

\subsection{What questions' answers can be improved after alignment if their answers changed recently?}
From \autoref{tab:target-year-error-analysis}, we notice that 829 questions cannot be answered correctly by the unaligned model but can be answered based on the knowledge in 2022 after target-year finetuning, i.e., finetuning improves the model's target-year performance $\mathcal{F}^{2022}$ on those questions. We show five of these questions for demonstration in \autoref{tab:finetune-improve}. 

\begin{table*}[]
\centering
\small
\renewcommand{\arraystretch}{1.4}
\renewcommand{\cellalign}{lc}
\begin{tabular}{@{}lll@{}}
\toprule
Question                                                                                                                      & Unaligned Answer  & Target-year Ft. Answer  \\ \midrule
\makecell{Which company is the shirt sponsor on the chest of\\ Chelsea F.C.'s kit?}              & Yokohama          & Three                   \\ \midrule
\makecell{Which club was the runner-up in the most recently held\\ UEFA Champions League final?} & Tottenham Hotspur & Liverpool               \\ \midrule
Who is the current Formula One World Drivers' Champion?                                                                       & Lewis Hamilton    & Max Verstappen          \\ \midrule
What film features Sam Neill in one of the leading roles?                                                                     & Jurassic Park     & Jurassic World Dominion \\ \midrule
What is the theme for Burning Man?                                                                                            & Metamorphoses     & Waking Dreams           \\ \bottomrule
\end{tabular}
\caption{Sampled questions that cannot be answered correctly by the unaligned model but can be answered with the correct answer as of 2022 after target-year finetuning.}
\label{tab:finetune-improve}
\end{table*}

From \autoref{fig:unaligned-correct-popularity}, we can see that popularity has a greater influence on the correctness of the target-year finetuning model than the unaligned one, which proves our finding in \S\ref{subsec:improvement_analysis} again that popular questions are easier to be temporally-aligned. Also, we notice from \autoref{tab:earliest-correct-year} that there are higher proportions of questions whose answers became true in 2022 than in 2021 and 2020, compared to the proportions that the unaligned LM answers correctly. This suggests that temporally-aligned LMs do not simply answer time-sensitive questions based on the occurrence of their answers in their pretraining corpora like the unaligned model. In contrast, they can learn the status of temporal knowledge through finetuning to answer more questions with the knowledge that has changed more recently.

\subsection{What kind of questions can LMs not answer correctly even if being temporally aligned?}
Some questions can still not be answered correctly before or after alignment. From \autoref{tab:target-year-error-analysis}, we see that 61.5\% (4070/6619) of the recent-answer-changed questions cannot be answered correctly by the target-year finetuning LM. We show some examples in \autoref{tab:finetune-incorrect}. In detail, for popular questions like ``the latest edition of Super Bowl'' or ``manager of Rangers FC'', the finetuned LM can respond with partially correct answers with different formats as the ground truth. This suggests one of our automatically generated TAQA dataset's limitations: some of the answers can be noisy or incomplete. As the answers are directly extracted from tables on Wikipedia, they might not contain enough contextual information for the questions, and oftentimes there are nationalities labeled with people's names in such tables. In the meantime, the finetuned LM also performs poorly on questions for numerical answers, as LMs cannot learn sufficient mathematical capabilities from either pretraining or finetuning. Moreover, for less-popular questions, finetuning's effect becomes poorer, and the aligned LM cannot even give partially correct answers.

\begin{table*}[]
\centering
\small
\renewcommand{\arraystretch}{1.4}
\renewcommand{\cellalign}{lc}
\begin{tabular}{@{}lll@{}}
\toprule
Question & Target-year Ft. Answer   & Ground truths  \\ \midrule
\makecell{What is the latest edition of the Super Bowl \\ game that was played?}  & Super Bowl LVI  & LVI   \\ \midrule
What are Walmart's total assets?   & \$260.84 billion          & 244    \\ \midrule
\makecell{What is the title of the show where Taraji P. Henson\\ appeared in her latest television role?} & The Best of the Oscars   & \makecell{That's My Jam, BET Awards,\\ RuPaul's Drag Race}  \\ \midrule
Who is the most recent manager of Rangers F.C.?   & Giovanni van Bronckhorst & \makecell{Netherlands Giovanni van Bronckhorst,\\ England Michael Beale} \\ \midrule
\makecell{Which team was the runner-up in the latest\\ WSE Cup Finals?}  & GKS Tychy    & HC Braga, Follonica Hockey  \\ 
\bottomrule
\end{tabular}
\caption{Sampled questions that cannot be answered correctly as of the knowledge in 2022 by the target-year finetuning model.}
\label{tab:finetune-incorrect}
\end{table*}

\subsection{What errors caused by target-year finetuning can be improved by the temporal-adaptive finetuning method?}
One of the shortcomings of the target-year finetuning method is that it might make errors when it gives answers that have never been true in history for time-sensitive questions, as shown in \autoref{tab:main-results} where the $\mathcal{F}_{\text{max}}$ metric of the finetuned-to-2022 LM is less than the unaligned model.

\begin{figure}[htbp]
    \centering
    \includegraphics[width=0.5\linewidth]{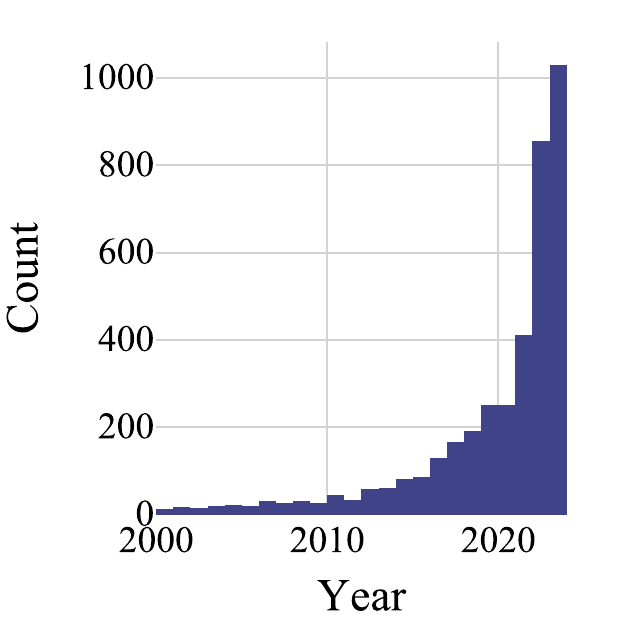}
    \caption{The distribution of the most recent year of the answers given by the temporal-adaptive LM to be correct.}
    \label{fig:most-recent-adaptive-year}
\end{figure}

In total, there are 3875 out of 6619 answer-changed questions that the temporal-adaptive finetuning LM can give answers that have been true in history, where 735 among them cannot be answered correctly by the target-year finetuning LM. Examples are shown in \autoref{tab:adaptive-recover}, where temporal-adaptive LMs can dynamically answer those questions with answers ranging from different times. We present the distribution of the most recent year that the temporal-adaptive finetuned LM's response to be correct in \autoref{fig:most-recent-adaptive-year}. In total, 59.2\% of questions can be answered based on knowledge after 2020, suggesting that there is still room for improvement to let temporal-adaptive LMs elicit more recent knowledge from finetuning. We leave this as one of the problems worth investigating in the future.

\begin{table*}[ht]
\centering
\small
\renewcommand{\arraystretch}{1.4}
\renewcommand{\cellalign}{lc}
\begin{tabular}{@{}lll@{}}
\toprule
Question  & Target-year Ft. Answer  & Temporal-adaptive Ft. Answer      \\ \midrule
\makecell{What is the title of the show where Taraji P. Henson\\ appeared in her latest television role?} & The Best of the Oscars                                                                & Empire (2015-2020)                \\ \midrule
Who is the current shirt sponsor of Sevilla FC?  & Codere   & Marathonbet (2019-2021)           \\ \midrule
\makecell{What is the most recent film that Matthew Lillard\\ has lent his voice to?}  & \makecell{Teenage Mutant Ninja\\ Turtles: Mutant Mayhem} & Trick or Treat Scooby-Doo! (2022) \\ \midrule
\makecell{In which city was the International Philosophy\\ Olympiad last held?}                           & Sarajevo   & Oslo (2021)   \\ \midrule
\makecell{Who is the captain of the winning team in the\\ last UEFA Champions League final?}              & Thiago  & Iker Casillas (2014) \\ 
\bottomrule
\end{tabular}
\caption{Sampled questions that cannot be answered correctly as of the knowledge in 2022 by the target-year finetuning model but can be responded with historically correct answers by the temporal-adaptive finetuned model. The years that the temporal-adaptive finetuned LM's answers being correct are shown in the parenthesis.}
\label{tab:adaptive-recover}
\end{table*}

\section{Detailed F1 scores of unaligned and aligned LMs}

To provide detailed information, we demonstrate the detailed target-year F1 scores of the LMs presented in \autoref{fig:teaser} regarding every year between 2000 to 2023 in \autoref{tab:knowledge-distribution}. 

\begin{sidewaystable}[htbp]
\centering
\scriptsize
\begin{tabular}{@{}lcccccccccccccccccccccccc@{}}
\toprule
Model $\downarrow$ Year $\rightarrow$ & 2000 & 2001 & 2002 & 2003 & 2004 & 2005 & 2006 & 2007 & 2008 & 2009 & 2010 & 2011 & 2012 & 2013 & 2014 & 2015 & 2016 & 2017 & 2018 & 2019 & 2020 & 2021 & 2022 & 2023 \\ \midrule
Davinci-002                          & 6.4  & 6.7  & 6.8  & 7.0  & 7.3  & 8.0  & 8.3  & 8.7  & 9.4  & 9.6  & 10.2 & 11.5 & 12.3 & 13.8 & 14.8 & 15.1 & 15.6 & \textbf{16.6} & 16.4 & 15.0 & 12.3 & 11.4 & 10.0 & 9.0  \\
LLaMA1-65B                           & 6.7  & 7.2  & 7.3  & 7.7  & 7.7  & 8.9  & 9.4  & 10.3 & 10.8 & 11.5 & 12.0 & 13.7 & 15.0 & 16.7 & 18.1 & 19.2 & 19.3 & 20.0 & \textbf{20.1} & 18.8 & 14.3 & 13.2 & 12.1 & 10.3 \\
LLaMA2-70B                           & 6.4  & 6.5  & 6.6  & 7.1  & 7.0  & 8.0  & 7.9  & 9.0  & 9.1  & 9.3  & 10.2 & 11.1 & 12.0 & 13.7 & 15.0 & 16.6 & 17.4 & 19.7 & 21.5 & \textbf{23.0} & 20.6 & 19.8 & 17.2 & 14.3 \\
LLaMA2-7B                            & 5.7  & 6.0  & 6.1  & 6.4  & 6.5  & 7.1  & 7.4  & 7.9  & 8.0  & 8.6  & 8.9  & 9.5  & 10.1 & 11.4 & 12.1 & 12.9 & 13.2 & 13.6 & 13.9 & \textbf{14.1} & 12.0 & 11.4 & 10.4 & 9.4  \\
ChatGPT                              & 5.2  & 5.4  & 5.2  & 5.5  & 5.6  & 5.9  & 6.1  & 6.6  & 6.9  & 7.1  & 7.6  & 8.1  & 8.6  & 9.9  & 10.5 & 11.8 & 12.5 & 14.1 & 15.8 & 19.4 & 23.0 & \textbf{23.8} & 17.1 & 14.2 \\
Tulu2                                & 5.6  & 5.9  & 5.9  & 6.4  & 6.4  & 7.0  & 7.3  & 7.8  & 8.0  & 8.1  & 8.6  & 9.3  & 10.0 & 11.4 & 12.7 & 13.6 & 14.5 & 16.3 & 18.7 & \textbf{22.9} & 22.5 & 19.1 & 16.2 & 13.5 \\
LLaMA2-70B Fine-tuned                & 6.2  & 6.6  & 6.7  & 7.1  & 7.3  & 7.8  & 8.1  & 8.3  & 8.5  & 8.6  & 9.3  & 9.9  & 10.4 & 11.6 & 12.3 & 13.6 & 14.3 & 15.5 & 16.9 & 19.1 & 19.8 & 23.7 & \textbf{27.9} & 19.1 \\
LLaMA2-70B Temporal adaptive         & 6.7  & 6.9  & 7.1  & 7.6  & 7.8  & 8.1  & 8.6  & 9.1  & 9.3  & 9.5  & 10.0 & 10.8 & 11.5 & 12.9 & 13.9 & 15.4 & 16.2 & 17.8 & 19.0 & 20.7 & 19.9 & 23.5 & \textbf{25.7} & 17.6 \\ \bottomrule
\end{tabular}
\caption{The target-year F1 score $\mathcal{F}^y$ regarding every year between 2000 and 2023 on models demonstrated in \autoref{fig:teaser}. The maximum F1 score of each LM is bolded.}
\label{tab:knowledge-distribution}
\end{sidewaystable}

\end{document}